%% file: main.tex
\documentclass[]{mila}

\title{Sampler Replay: Train The Sampler Not The Trainer \\
in Reinforcement Learning with Verifiable Rewards}
\author[1,*]{Milad Aghajohari}
\author[1,4,*]{Kamran Chitsaz}
\author[1,3,4,5,\dagger]{\newline Sarath Chandar}
\author[1,\dagger]{Alessandro Sordoni}
\author[1,2\dagger]{Michael Noukhovitch}
\author[1,2,5,\dagger]{Aaron Courville}

\affiliation[1]{Mila}
\affiliation[2]{Universit\'e de Montr\'eal}
\affiliation[3]{Polytechnique Montr\'eal}
\affiliation[4]{Chandar Research Lab}
\affiliation[5]{Canada CIFAR AI Chair}

\contribution[*]{Equal contribution}
\contribution[\dagger]{Equal advising\renewcommand{\thefootnote}{}\footnote{Correspondence: \texttt{\{aghajohm,kamran.chitsaz\}@mila.quebec} }}

\usepackage[utf8]{inputenc} %
\usepackage[T1]{fontenc}    %
\usepackage{url}            %
\usepackage{booktabs}       %
\usepackage{amsfonts}       %
\usepackage{nicefrac}       %
\usepackage{microtype}      %
\usepackage{xcolor}         %
\usepackage{minted}
\usepackage{caption}
\input{math_commands.tex}

\usepackage{algorithm}
\usepackage[noend]{algpseudocode}
\usepackage{xcolor}
\usepackage{booktabs}
\usepackage{threeparttable}
\definecolor{darkGreen}{rgb}{0.2,0.5,0.2}
\definecolor{mydarkblue}{rgb}{0,0.08,0.45}
\usepackage{url}
\usepackage{graphicx}
\usepackage{subcaption}
\usepackage{enumitem}
\usepackage{booktabs,threeparttable}
\usepackage{tikz}           %
\usepackage{multirow}
\usepackage{mathtools}
\usepackage{etoc}
\usepackage{wrapfig}
\usepackage{dsfont}

\usetikzlibrary{patterns,patterns.meta,positioning,backgrounds,calc}

\input{assets/figstyles}

\usepackage{amsthm}
\usepackage[colorlinks]{hyperref}
\usepackage[capitalize]{cleveref}
\usepackage{todonotes}
\theoremstyle{definition}
\makeatletter
\@ifundefined{theorem}{\newtheorem{theorem}{Theorem}}{}
\makeatother
\definecolor{thmbg}{HTML}{FAF6F8} 
\tcolorboxenvironment{theorem}{enhanced, frame hidden, colback=thmbg, arc=2pt, left=10pt, right=10pt, top=5pt, bottom=5pt}

\abstract{
Train-inference mismatch is a well-known problem that degrades reinforcement learning with verifiable rewards (RLVR) performance. This problem is currently framed as training the trainer with samples coming from an imperfect version of itself: a sampler optimized to meet some efficiency criteria. Therefore, current RLVR algorithms are designed to account for the discrepancy in the trainer's gradient due to the sampler's off-policy rollouts. In this paper, we reframe the problem: we argue that RLVR should strive to \emph{train the sampler}, since it is the sampler's performance at test time that we ultimately care about. Therefore, we view the trainer merely as a mechanism to get an approximation of the sampler's true gradient. We propose \emph{Sampler Replay}, an algorithm that replays the sampler's hidden states, logits, and routing decisions in the trainer's forward pass to better approximate the sampler's gradient. Across tasks and model scales, we show that Sampler Replay stabilizes training under severe train-inference mismatch induced by quantization (weights, activations, KV cache), weight perturbation, and staleness, outperforming strong baselines. Furthermore, we show that Sampler Replay is a generalization of techniques that have been widely used in the literature to address TIM but that, so far, have been treated independently. Notably, we mathematically prove that Score Centering is a special case of Sampler Replay, where only the logits are replayed. 
}

\begin{document}

\maketitle

{\renewcommand{\thefootnote}{}\footnotetext{Code repository: \url{https://github.com/chandar-lab/Sampler-Replay}}

\section{Introduction}
Reinforcement Learning with Verifiable Rewards (RLVR) is a key part of LLM pipelines but has a fundamental issue affecting training stability: Train-Inference Mismatch (TIM). The policy that generates the rollouts, the sampler, is different from the policy that estimates the gradient, the trainer. The optimally efficient way to train fundamentally differs from the optimally efficient way to do inference leading to different engines, kernels, batch sizes, quantization, and parameter staleness. TIM can cause training instability and reduces RLVR learning efficiency \citep{qi2025defeating,liu-li-2025-rl-collapse,he2025nondeterminism,yao2025offpolicy}. 

Current solutions to TIM treat the trainer as the target of training process and the sampler merely the means to that end. From this perspective, gradient estimates derived from samples are biased (off-policy) with respect to the the trainer's policy and should be corrected through importance sampling  \citep{yao2025offpolicy,liu-li-2025-rl-collapse,zheng2025gspo,lingteam2025ring1t_icepop,chen2025minimax}. We argue that an effective sampler -- served as an inference model -- is the true objective of training. We frame TIM as a problem of how an off-policy trainer can give effective gradients for a non-differentiable sampler. In short, how can we leverage RL to effectively train the sampler, not the trainer?

To align the trainer's gradient with the sampler, we propose \emph{Sampler Replay}, leveraging the states of the sampler in the trainer's forward pass. Sampler Replay is a paradigm that saves sampler's forward pass state and forcibly replays it during the trainer's forward pass to better align the gradient, as shown in \autoref{fig:sampler-replay}.

The Sampler Replay paradigm explains the effectiveness and generalizes the concept behind many state-of-the-art TIM methods. Replaying the sampler's top-p choices, known as Keep Sampling Mask \citep{deepseek2025v32,vllm2026samplingmask}, and replaying the sampler's expert routing decisions \citep{ma2025r3} are straightforward examples of Sampler Replay. Furthermore, we show that Score Centering \citep{marek2026score}, a recent RLVR stabilization method, is mathematically equivalent to replaying the sampler's logits, making it another subset of Sampler Replay. We generalize these approaches by including hidden state replay throughout the model. By replaying hidden states, as well as logits, we demonstrate improved RL stability, and better final sampler performance across a variety of setups. We summarize our contributions:

\begin{enumerate}[leftmargin=*, labelsep=0.5em]
    \item \textbf{We reframe train-inference mismatch in RLVR.}
    Rather than viewing the sampler as an imperfect rollout generator
    for the trainer, we view the trainer as an imperfect gradient
    estimator for the sampler.

    \item \textbf{We propose Sampler Replay, a paradigm for effectively training the sampler encompassing several
    RLVR stabilization methods.}
    We prove that Score Centering is mathematically
    equivalent to replaying the sampler's logits, making it a
    special case of Sampler Replay. We generalize the approach and propose Hidden State Replay to replay also the hidden states of the sampler. 

    \item{\textbf{We propose Activation Trust Weighting (ATW), an alternative to Truncated Importance Sampling (TIS) for Sampler Replay.}}
    TIS is frequently added to account for both TIM and data staleness. We find that it can conflict with Sampler Replay and frequently harms Score Centering performance. We propose ATW, a novel activation-space reweighting to reduce harmful gradient updates.

    \item 
    \textbf{We demonstrate that Sampler Replay + ATW improves RL stability and performance at scale.}
    By replaying hidden states, logits and expert choices, we show higher stability and performance across a variety of 
    mismatches (quantization, weight perturbation, and staleness),
    tasks, and model scales. We show that ATW further increases training stability, especially in the presence of staleness.

\end{enumerate}

\begin{figure}[t]
\centering
\includegraphics[width=0.75\linewidth]{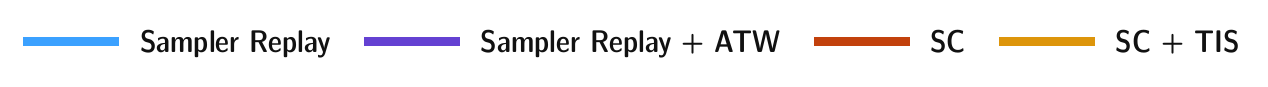}\par\vspace{-5pt}
\newsavebox{\srteaser}\sbox{\srteaser}{\def\srreplayy{1.3cm}\input{sampler_replay}}
\resizebox{\linewidth}{!}{$\vcenter{\hbox{\usebox{\srteaser}}}$\hspace{0.3cm}$\vcenter{\hbox{\includegraphics[height=\dimexpr\ht\srteaser+\dp\srteaser\relax]{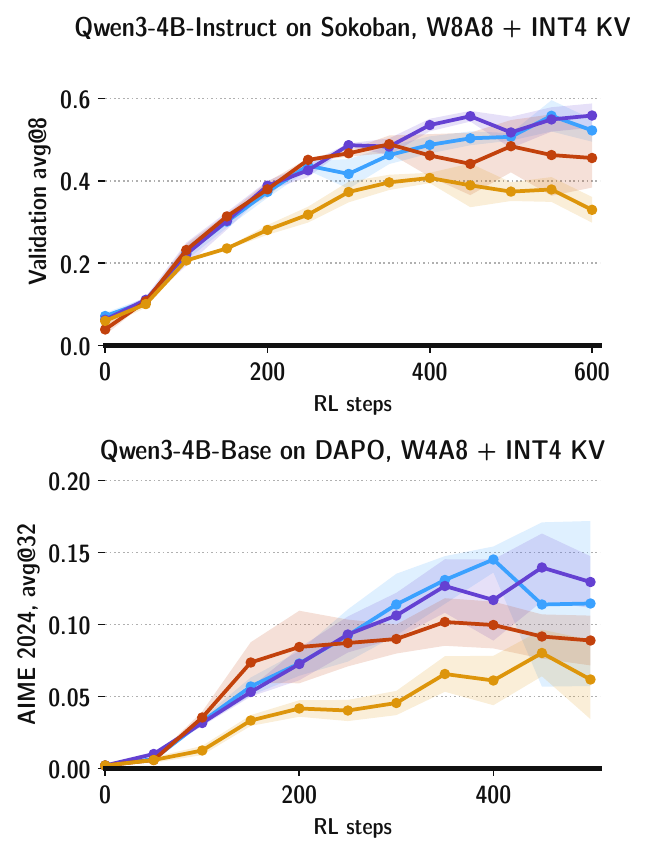}}}$}
\caption{Left: Sampler replay. The sampler's hidden states and its logits are replayed in the trainer's forward pass. Right top: Qwen3-4B-Instruct-2507 on Sokoban with a W8A8 + INT4 KV sampler. Both Sampler Replay and Sampler Replay + ATW ($\tau=0.1$) are stable and achieve higher reward than SC and SC+TIS. Right bottom: Qwen3-4B-Base on DAPO with a W4A8 + INT4 KV sampler. Sampler Replay + ATW ($\tau=0.3$) is the most stable and performs the best. SC+TIS trains slower than SC in both.}
\label{fig:sampler-replay}
\end{figure}

\newcommand{\grad}{\operatorname{\nabla}}
\newcommand{\gradtheta}{\operatorname{\nabla_{\theta}}}
\newcommand{\gradphi}{\operatorname{\nabla_{\phi}}}
\newcommand{\sg}{\operatorname{sg}}
\newcommand{\logpi}{\log \pi(y|x)}
\newcommand{\logphi}{\log \phi(y|x)}

\section{Train the Sampler, Not the Trainer}

RLVR typically uses two engines: an inference engine to generate rollouts and a training engine to compute gradients. Let $\mu$ denote the sampler policy with parameters $\phi$, and $\pi$ the trainer policy with parameters $\theta$ ($\theta = \phi$ in synchronous setups). Train-inference mismatch occurs when $\mu \neq \pi$. TIM is ubiquitous. It arises from differences in CUDA kernels, floating-point arithmetic, quantization, and weight staleness. Even mathematically equivalent operations can produce different results due to floating-point non-associativity. These differences propagate through layers, affecting intermediate hidden states and ultimately the next-token distribution. Existing work on RLVR with TIM treats the trainer as the policy to improve. We instead seek to improve the sampler. This shift changes how we explain TIM and design algorithms to address it.

\subsection{Training the Trainer}
The trainer $\pi$ is the target of RLVR training. To improve the trainer, we estimate the gradient of trainer's expected reward
\begin{equation}
\label{eq:gpg}
g_\pi = \gradtheta \mathbb{E}_{x \sim D, y \sim \pi} R(x, y) = \mathbb{E}_{x\sim D, y \sim \pi} R(x,y) \gradtheta \logpi
\end{equation}

This expectation cannot be estimated directly because only $\mu$ generates rollouts. Substituting $\mu$ results in the following biased gradient estimate
\begin{equation}
\label{eq:gpg_mu}
\mathbb{E}_{x\sim D, y \sim \mu} R(x,y) \gradtheta \logpi
\end{equation}
In this view, TIM degrades RLVR because the rollouts are off-policy. The goal is therefore to correct this mismatch.
Applying an importance ratio $\frac{\pi(y)}{\mu(y)}$ corrects this mismatch.

\begin{equation}
g_\pi = \gradtheta \mathbb{E}_{x \sim D, y \sim \mu} \frac{\pi(y)}{\mu(y)} R(x,y) = \mathbb{E}_{x \sim D, y \sim \mu} \frac{\pi(y)}{\mu(y)} R(x,y) \gradtheta \logpi.
\end{equation}

The importance ratio is unbounded and can induce high variance. Most RLVR algorithms mitigate this issue by applying
importance sampling at the token level, ignoring the
prefix distribution mismatch, and further trading off
bias for variance reduction through masking
\citep{liu-li-2025-rl-collapse, lingteam2025ring1t_icepop,
qi2026stablerl}, truncation \citep{yao2025offpolicy,
chen2025minimax}, or clipping \citep{schulman2017ppo}.

\subsection{Training the Sampler}
The sampler $\mu$ is the target of training. 
We improve the sampler by estimating the sampler's gradient of expected rewards

\begin{equation}
g_\mu = \gradphi \mathbb{E}_{x\sim D,y\sim\mu}R(x, y) = \mathbb{E}_{x\sim D,y\sim\mu} R(x, y) \gradphi \log \mu(y|x).
\end{equation}

However, the inference engine does not support backpropagation. We substitute the trainer's gradient, $\gradtheta \logpi$ as a surrogate for $\gradphi \log \mu(y|x)$ and use $\mathbb{E}_{x\sim D,y\sim\mu} R(x, y) \gradtheta \logpi$ to update the sampler. We arrive at the same estimator as in \Cref{eq:gpg_mu}, but for a different reason.

If $\mu \approx \pi$, this surrogate may approximate $g_\mu$ well. From this perspective, TIM degrades RLVR by making this surrogate less reliable. Improving the alignment between the surrogate gradient and $g_\mu$ thus offers another way to mitigate TIM. This perspective motivates Sampler Replay.

\section{Sampler Replay Algorithm}
Sampler Replay aligns the trainer's forward pass, when constructing the computational graph of the loss, with the sampler's recorded forward pass. Since the backward pass depends on the values encountered during the forward pass, we expect this intervention to yield a better approximation of the sampler's backward pass. Sampler Replay replays the hidden states, logits, and router of the sampler in the trainer, preventing the accumulation of computational differences that cause the forward passes to diverge. 

Formally, let $\mathcal{F}(x)$ be an operation in the architecture that receives input $x$ and outputs $y$. Let $\mathcal{F}_\pi$ and $\mathcal{F}_\mu$ refer to trainer and sampler operations, respectively. Replaying here means replacing $y = \mathcal{F}_\pi(x)$ with $\mathcal{F}_\pi(x) +\sg(\mathcal{F}_\mu(x)-\mathcal{F}_\pi(x))$.  This way, the sampler's computation gets replayed along the forward pass preventing the divergence, while the backward pass computes gradient through the trainer. This trick is known as Straight-Through-Estimator (STE). We denote by $\pi_\theta^{\mathrm{replay}}$ the trainer's computational graph with the sampler's recorded values replayed. To make it more concrete, we define how we replay each.

\textbf{Hidden States Replay} We record the hidden states of the sampler in inference time. In training time, we replace the trainer's hidden states at layer $i$ with $h^i_\pi = h^i_\pi+\sg(h^i_\mu-h^i_\pi)$.

\textbf{Logits Replay} We record the logits of the sampler for the vocabulary tokens in inference time. In training time, we replace trainer's logits by $l_\pi = l_\pi+sg(l_\mu-l_\pi)$. We prove that Logits Replay is mathematically equivalent to Score Centering in \Cref{sec:score_centering_equals_logits_replay}.

\textbf{Rollout Router Replay (R3)}
In Mixture-of-Expert models, we also replay the router following \citet{ma2025r3}. 

\subsection{Score Centering is Mathematically Equivalent to Replaying Sampler Logits}
\label{sec:score_centering_equals_logits_replay}
At a fixed prefix, the trainer's \emph{score} for a token is the gradient of that token's log-probability with respect to the trainer's parameters. These scores average to zero under the trainer's next-token distribution and can have a non-zero mean under the sampler's distribution. The distribution mismatch introduces a “drift” term in the gradient expression which entails that the trainer’s expected update might be non-zero even for a constant reward. This drift term disappears when doing importance sampling at the cost of high variance. Instead of using importance sampling, Score Centering is a recent method that proved high stability and addresses the drift by removing the sampler-weighted mean directly from the score~\citep{marek2026score}. We prove that full-vocabulary Score Centering is a special case of Sampler Replay.

\begin{theorem}[Score Centering is Replaying Sampler Logits]
Let $p=\operatorname{softmax}(z_{\mathrm{trainer}})$ and $q=\operatorname{softmax}(z_{\mathrm{sampler}})$ be the trainer’s and the sampler’s next-token distributions at a fixed prefix. For each token $v$ in vocabulary $\mathcal V$, let $p_v$ and $q_v$ denote the probabilities that the trainer and sampler assign to $v$, respectively. Let $y$ be the sampled token with advantage $A$. Compare full-vocabulary Score Centering,
\[
\mathcal{L}_{\mathrm{SC}}=-A\left(\log p_y-\sum_v q_v\log p_v\right),
\] 
with policy-gradient loss when replaying sampler logits,
\[
\mathcal{L}_{\mathrm{replay}}=-A\log\operatorname{softmax}(\tilde z)_y, \qquad \tilde z=z_{\mathrm{trainer}}+\operatorname{stopgrad}\left(z_{\mathrm{sampler}}-z_{\mathrm{trainer}}\right).
\]
The two losses have the same gradient:
\[
\nabla_{z_{\mathrm{trainer}}}\mathcal{L}_{\mathrm{SC}}=\nabla_{z_{\mathrm{trainer}}}\mathcal{L}_{\mathrm{replay}}=-A(e_y-q).
\]
where $e_y$ denotes $y$'s one-hot vector.
\end{theorem}

\begin{proof}
For any logit vector $z$ and token $v$, differentiating the log-softmax gives
\[
\frac{\partial}{\partial z_u}\log[\operatorname{softmax}(z)]_v
=
\frac{\partial}{\partial z_u}
\left(z_v-\log\sum_{w\in\mathcal V}e^{z_w}\right)
=
\mathds{1}_{\{u=v\}}-[\operatorname{softmax}(z)]_u.
\]
In vector form, this is
$\nabla_z\log[\operatorname{softmax}(z)]_v=e_v-\operatorname{softmax}(z)$,
where $e_v$ is the one-hot vector for token $v$.

Applying this identity, we differentiate the Score Centering loss
\[
\nabla_{z_{\mathrm{trainer}}}\mathcal L_{\mathrm{SC}}
=-A\left[(e_y-p)-\sum_{v\in\mathcal V}q_v(e_v-p)\right].
\]

We expand the sum as

\begin{equation}
\sum_{v\in\mathcal V}q_v(e_v-p)
=\sum_{v\in\mathcal V}q_v e_v-\sum_{v\in\mathcal V}q_v p
=\underbrace{\sum_{v\in\mathcal V}q_v e_v}_{q}
  -\underbrace{\left(\sum_{v\in\mathcal V}q_v\right)}_{1}p
=q-p.
\end{equation}

The first sum equals $q$ because $q_v e_v$ contributes $q_v$ to
coordinate $v$ and zero to every other coordinate. The scalar sum
equals one because $q$ is a probability distribution.

Substituting gives
\[
\nabla_{z_{\mathrm{trainer}}}\mathcal L_{\mathrm{SC}}
=-A\bigl[(e_y-p)-(q-p)\bigr]
=-A(e_y-q).
\]

For sampler replay, applying the chain rule through $\tilde z$ gives
\[
\begin{aligned}
\nabla_{z_{\mathrm{trainer}}}\mathcal L_{\mathrm{replay}}
&=\left(\frac{\partial\tilde z}{\partial z_{\mathrm{trainer}}}\right)^{\!\top}
  \nabla_{\tilde z}\mathcal L_{\mathrm{replay}} \\
&=-A\left(\frac{\partial\tilde z}{\partial z_{\mathrm{trainer}}}\right)^{\!\top}
  \bigl(e_y-\operatorname{softmax}(\tilde z)\bigr),
\end{aligned}
\]
where the second line uses the same log-softmax derivative. In the
backward pass, the $z_{\mathrm{trainer}}$ term in $\tilde z$
contributes the identity and the stop-gradient term contributes zero,
so $\partial\tilde z/\partial z_{\mathrm{trainer}}=I$. In the forward
pass, $\tilde z=z_{\mathrm{sampler}}$, so
$\operatorname{softmax}(\tilde z)=q$. Substituting both into the
chain rule gives
\[
\nabla_{z_{\mathrm{trainer}}}\mathcal L_{\mathrm{replay}}
=-A(e_y-q)
=\nabla_{z_{\mathrm{trainer}}}\mathcal L_{\mathrm{SC}}.
\]
\end{proof}

\subsection{Activation Trust Weighting}
\label{sec:atw}

Sampler Replay attempts to reproduce the sampler's forward pass in the trainer, but they can still differ significantly for some tokens. 
For example, we find high TIM when the data was generated by a sampler many updates prior to it being passed to the trainer, i.e., data staleness. 
In prior work, TIS \citep{yao2025offpolicy} has been used to down-weight the gradient contribution of high-TIM tokens, but empirically we find it ineffective for Replay, see \Cref{app:atw_vs_tis}.
We therefore propose a novel heuristic to down-weight the gradient contributions of tokens. For each token, we measure how trustworthy is the trainer's approximation of the sampler's forward/backwards pass.
We use the agreement between the trainer's and the sampler's 
hidden state activations $h$ as a proxy for the reliability of the trainer's 
gradient estimate. Our intuition is that a large mismatch at 
even one layer $h^i$ can affect the gradients propagated through that 
layer to others. To conservatively measure the mismatch across all layers for some token $t$, we take the largest cosine distance between hidden states $h_i,t$ across all layers $T$:
\begin{equation}
    d_t = \max_{i \in T} 1-\cos(h^{\mu}_{i,t}, h^{\pi}_{i,t}).
\end{equation}
This measure ranges from zero when the hidden states agree in 
direction at every layer to two when they point in opposite 
directions at any layer. We hypothesize that the larger this mismatch, the less reliable 
the trainer's gradient estimate is. We therefore down-weight 
each token's gradient contribution linearly with its mismatch, 
using the following weight:
\begin{equation}
    w_t = \max\left(0, 1-\frac{d_t}{\tau}\right).
\end{equation}
This gives full weight to tokens with zero mismatch, progressively less weight to tokens with larger mismatch, and zero weight at and beyond a threshold $\tau > 0$. We call this \emph{Activation Trust Weighting} (ATW). Combining replay and weighting, we estimate the 
sampler's policy gradient as
\begin{equation}
\widehat{g}_\mu
=
\mathbb{E}_{x\sim D,\,y\sim\mu}
\left[
\sum_{t=1}^{|y|} A_t
\underbrace{
    \max\left(0,1-\frac{d_t}{\tau}\right)
}_{\text{
Trust Weighting}}
\underbrace{
    \vphantom{\max\left(0,1-\frac{d_t}{\tau}\right)}
    \nabla_\theta \log
    \pi_\theta^{\mathrm{replay}}(y_t\mid x,y_{<t})
}_{\text{Sampler replay}}
\right].
\end{equation}

\section{Experiments}
We empirically compare Sampler Replay, with and without Activation Trust Weighting, to Score Centering \citep{marek2026score}, with and without Truncated Importance Sampling \citep{yao2025offpolicy}, as the current strongest baselines across setups with high TIM. First, we reproduce experiments from \citet{marek2026score} and run Qwen3-0.6B on Countdow, inducing TIM through weight perturbations and quantization. Next, we focus on larger, more realistic setups where TIM arises from quantizing weights and the KV cache to improve inference throughput. We train Qwen3-0.6B-Base on GSM8K, Qwen3-4B-Base on DAPO, and Qwen3-4B-Instruct-2507 on Sokoban with 4-bit KV cache using standard quantization tricks (see \Cref{sec:sharp_bits_of_implementation}). To isolate TIM from data staleness, we run all previous setups in synchronous RL. We then explicitly test Sampler Replay with TIM due to data staleness, where the sampler lags behind the trainer, by training Qwen3-0.6B on Countdown and Qwen3.5-0.8B-Base on GSM8K. Finally, we train a Mixture-of-Experts Qwen3-30B-A3B-Base on INTELLECT-2 math, following \citet{marek2026score}. For each run, we report mean and standard error across 3 seeds, see \Cref{app:sec:hyperparameters} for full hyperparameters.

\subsection{Qwen3-0.6B on Countdown and Qwen3-0.6B-Base on GSM8K}
\begin{figure}[t]
\centering
\includegraphics[scale=0.47]{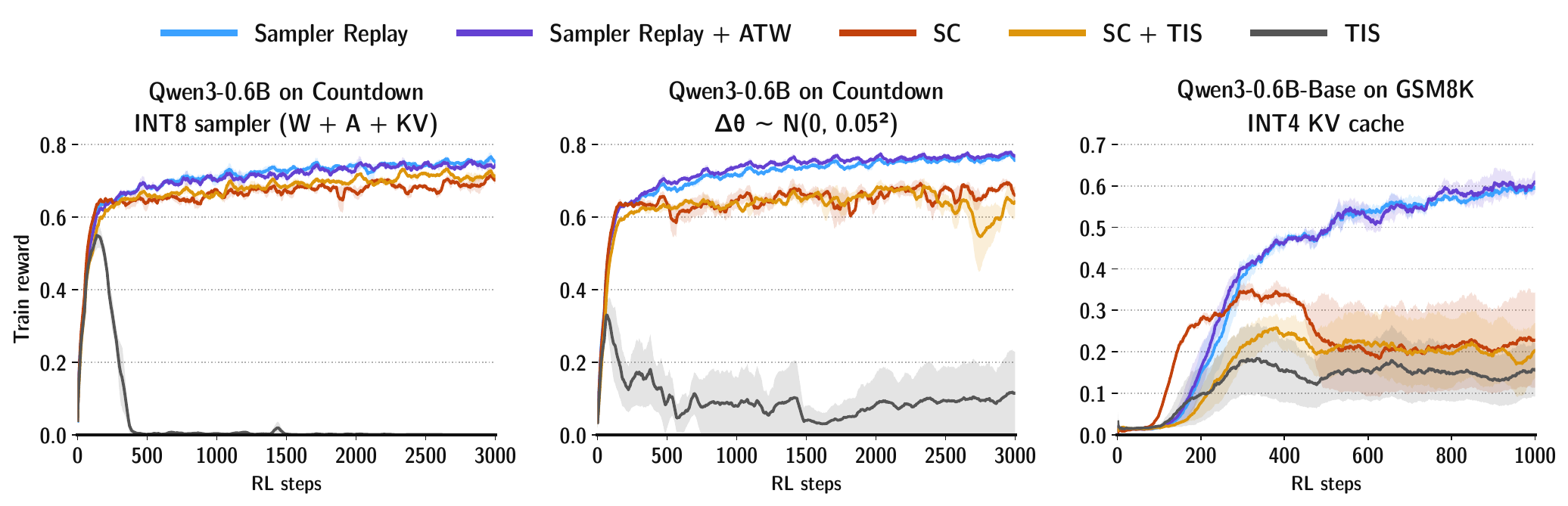}
\caption{Qwen3-0.6B on Countdown with an INT8 sampler and fixed noise on sampler, and Qwen3-0.6B-Base on GSM8K with an INT4 KV cache. Sampler Replay is more stable and achieves the highest reward. }
\label{fig:small_scale}
\end{figure}

\label{sec:small_scale}
First, we gain intuition about Sampler Replay through two small-scale tasks. We train Qwen3-0.6B on Countdown \citep{gandhi_stream_2024} with INT8 quantization. We quantize the KV cache, the weight matrices (including the token embedding and LM head) per output channel, and activations per token. As shown in \Cref{fig:small_scale} (left) Sampler Replay consistently achieves higher training reward than SC. SC+TIS improves on SC, but still underperforms Sampler Replay with or without ATW. 

Next, we create controlled TIM through weight perturbation. Following \citet{marek2026score}, we sample an offset at the start of training and add it to the sampler's copy of trainer weights, creating a fixed TIM. In \Cref{fig:small_scale} (middle), SC stalls at 0.65 and training becomes incredibly noisy as the trainer tries to account for an off-policy sampler. In contrast, Sampler Replay is minimally affected because its gradient focused on improving the sampler, ignoring any fixed TIM differences.  


We now extend to even more extreme TIM, with bf16 weights but a quantized INT4 KV cache, using Qwen3-0.6B-Base on GSM8K. 
As shown in \Cref{fig:small_scale} (right), SC and SC+TIS initially learn but quickly collapse. In contrast, Sampler Replay remains stable and continues to improve.

\begin{figure}[t]
\vspace{-5em}
\centering
\includegraphics[scale=0.47]{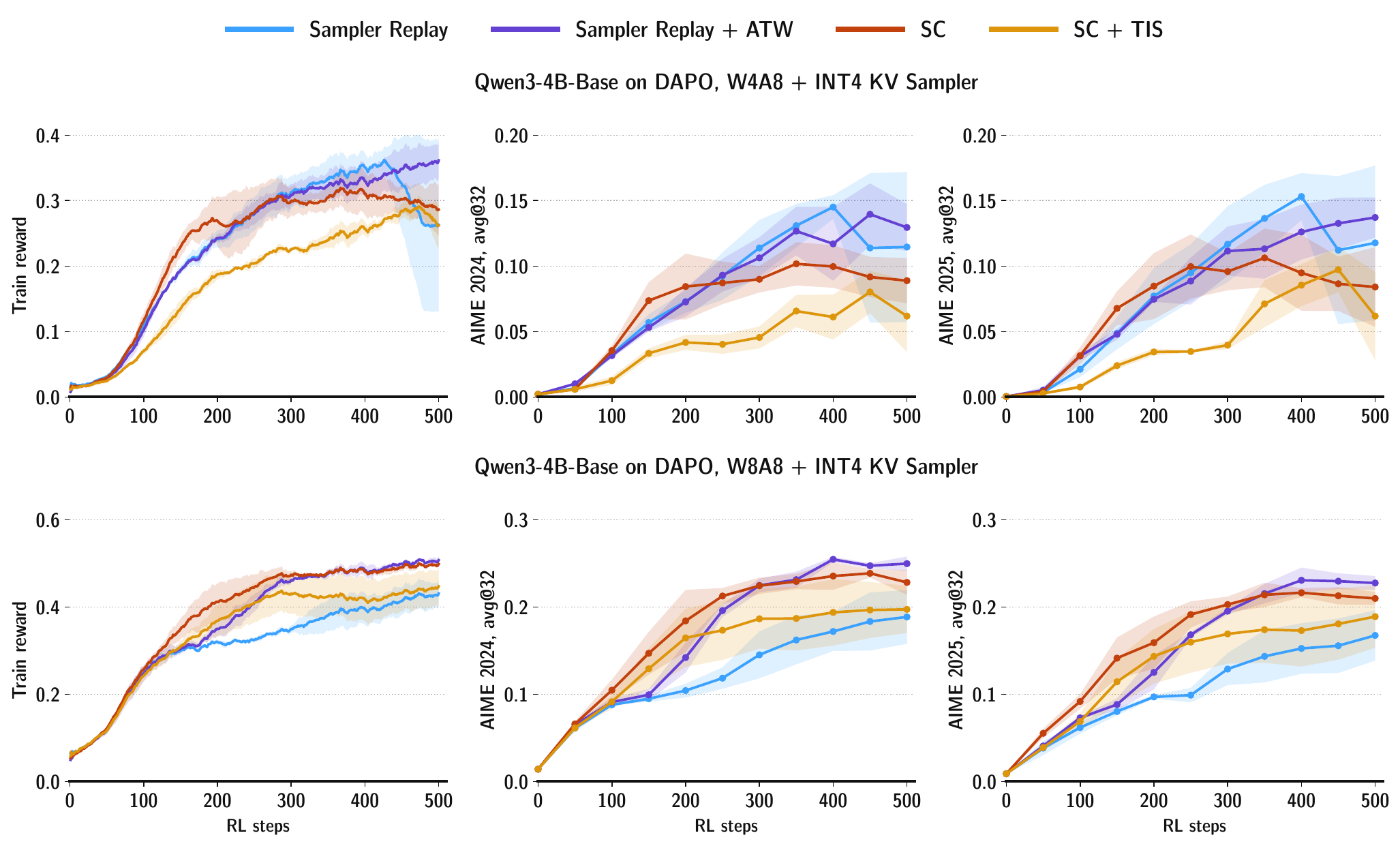}
\caption{Qwen3-4B-Base on DAPO with a W4A8 + INT4 KV sampler (top) and a W8A8 + INT4 KV sampler (bottom). Sampler Replay + ATW ($\tau=0.3$) is the most stable and performs the best. SC+TIS trains slower than SC. }
\label{fig:dapo}
\end{figure}

\subsection{Qwen3-4B-Base on DAPO}
Next, we scale to larger models and harder tasks. We train Qwen3-4B-Base on the English subset of DAPO-Math-17K \citep{yu_dapo_2025} and evaluate on AIME 2024 and 2025 math contests \citep{dekoninck_beyond_2026}. To test TIM, we quantize matrix-multiplications to either (4-bit - W4A8) or (8-bits - W8A8), and KV-cache to 4-bits. As shown in \Cref{fig:dapo}, Sampler Replay+ATW achieves the best performance across both tasks. Sampler Replay alone outperforms SC in higher TIM W4A8 setup, but underperforms in the more stable W8A8.
This shows the need for ATW to further close the TIM gap.
Interestingly, SC+TIS learns slower than SC and underperforms in both setups. This suggests that TIS can conflict with SC, motivating alternatives such as ATW which we explore further in \Cref{sec:sc_match}.

\begin{figure}[t]
\centering
\includegraphics[scale=0.57]{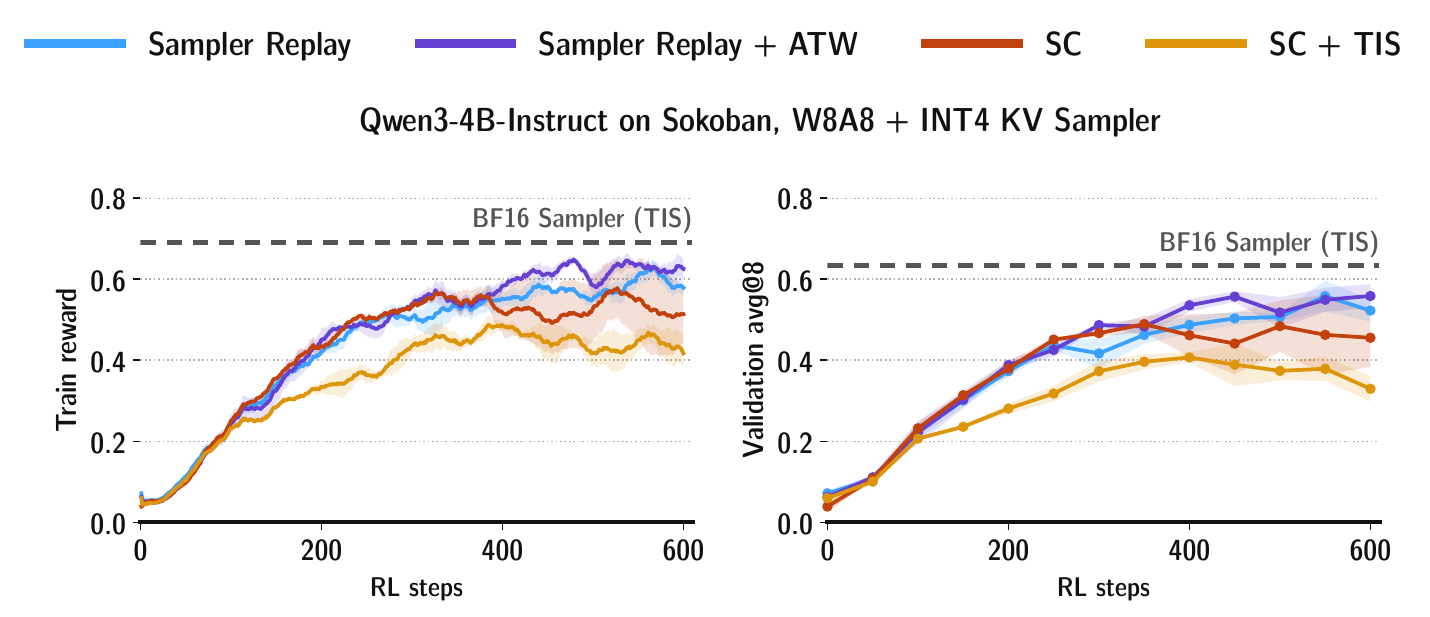}
\caption{Qwen3-4B-Instruct-2507 on Sokoban with a W8A8 + INT4 KV sampler. Both Sampler Replay and Sampler Replay + ATW ($\tau=0.1$) are stable and achieve higher reward than SC and SC+TIS. SC+TIS climbs slower than SC. }
\label{fig:sokoban}
\end{figure}

\subsection{Qwen3-4B-Instruct on Sokoban}
\label{sec:sokoban}

To evaluate on a complex OOD task, we train Qwen3-4B-Instruct-2507 on Sokoban puzzles. 
In the game, an agent plan a complex series of actions in order to push 3 boxes onto their target in a maze, given physical constraints.
These synthetically-generated puzzles greatly reduce the chance of eval contamination \citep{shao_spurious_2025} and mid-training exposure. 
We quantize the sampler with W8A8+KV-4 schema which reduces KV cache size four times. As shown in \Cref{fig:sokoban} Replay+ATW and Replay both achieve higher reward than both SC and SC+TIS. They are also more stable across seeds as confirmed by error bands. At step 600, Replay+ATW scores $55.9\%$ on validation while SC scores $45.6\%$. Additionally, SC+TIS trains worse than SC, showing that off-policy view is not suitable here. As a sanity check, we also train an unquantized BF16 sampler with TIS.

\subsection{Qwen3-30B-A3B-Base on INTELLECT-2 Math}
\label{sec:moe}
\begin{wrapfigure}[12]{r}{0.47\linewidth}
\vspace{-6em}
\centering
\includegraphics[scale=0.43]{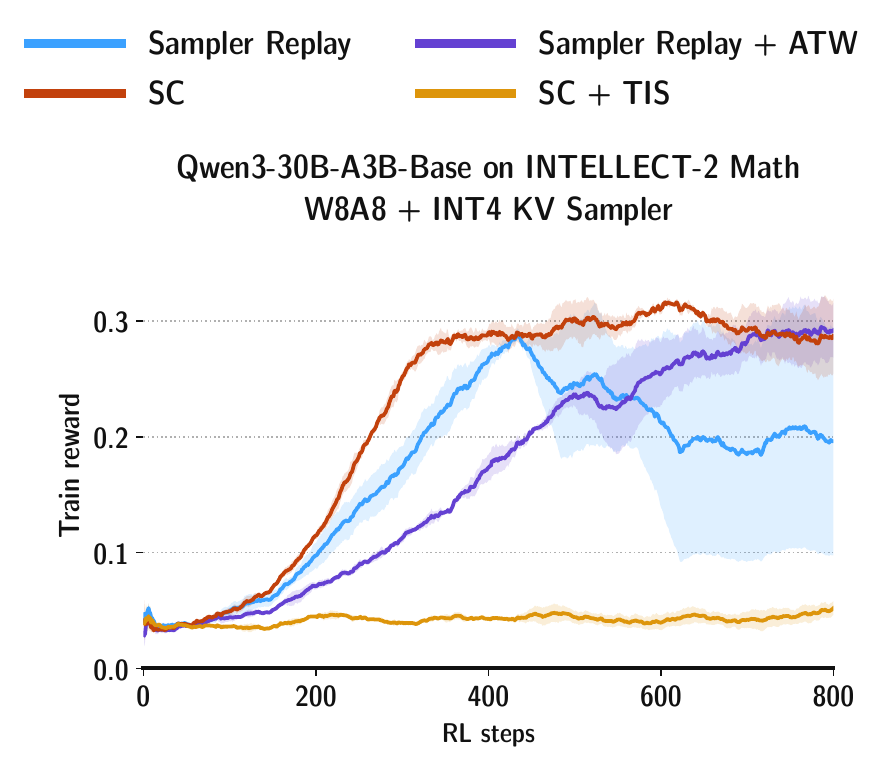}
\caption{Qwen3-30B-A3B-Base on INTELLECT-2 math, W8A8 + INT4 KV sampler; Sampler Replay+ATW ($\tau=0.3$) reaches SC's peak reward. SC+TIS stalls.}
\label{fig:moe}
\vspace{-1em}
\end{wrapfigure}
We now focus on the extra TIM of Mixture-of-Experts models and follow \citet{marek2026score} to train Qwen3-30B-A3B-Base on INTELLECT-2 math. We train with W8A8 weights and INT4 KV cache, SGD with learning rate of 0.01, a very small maximum context length of 1024 tokens, and a batch of 16 prompts with 8 responses per prompt. As shown in \Cref{fig:moe} Sampler Replay+ATW and SC achieve the same final performance, though SC achieves it earlier. We observe that ATW is crucial to stabilize Sampler Replay and prevent collapse. Notably, SC+TIS does not improve at all, demonstrating that TIS might harm SC performance. This setup, however, is somewhat unrealistic. First, following \citet{marek2026score} exactly we quantize KV without standard stabilization tricks (see \Cref{sec:sharp_bits_of_implementation}). This drops base model performance from $37\%$ to $4\%$ which does not recover even after $800$ training steps. Second, the $1024$ token context length is greatly insufficient for this base model.

\subsection{Staleness}
Next, we study TIM due to data staleness caused by multiple updates between sampling of the data and training on the data. We follow the minibatch synchronous RL setup of \citet{marek2026score,liu2026driftdistance}, where the sampler generates 64 minibatches at a time. The trainer then does 64 updates, making the data 0-63 steps stale. 
We train Qwen3-0.6B on Countdown 
and Qwen3.5-0.8B-Base on GSM8K. 
As shown in \Cref{fig:staleness} (left), Sampler Replay+ATW and SC+TIS succeed on Countdown. But neither Sampler Replay nor SC works without ATW or TIS, showing that these additions are necessary to account for staleness-induced TIM. On GSM8K shown in \Cref{fig:staleness} (right), only Sampler Replay+ATW improves reward. These setups might be confounded. TIS and ATW down-weight gradients effectively reducing the learning rate under SGD. This slows trainer moving away from the sampler. 
We discuss limitations of ATW under staleness in \Cref{sec:limitations}.
\label{sec:staleness}
\begin{figure}[h]
\centering
\includegraphics[scale=0.57]{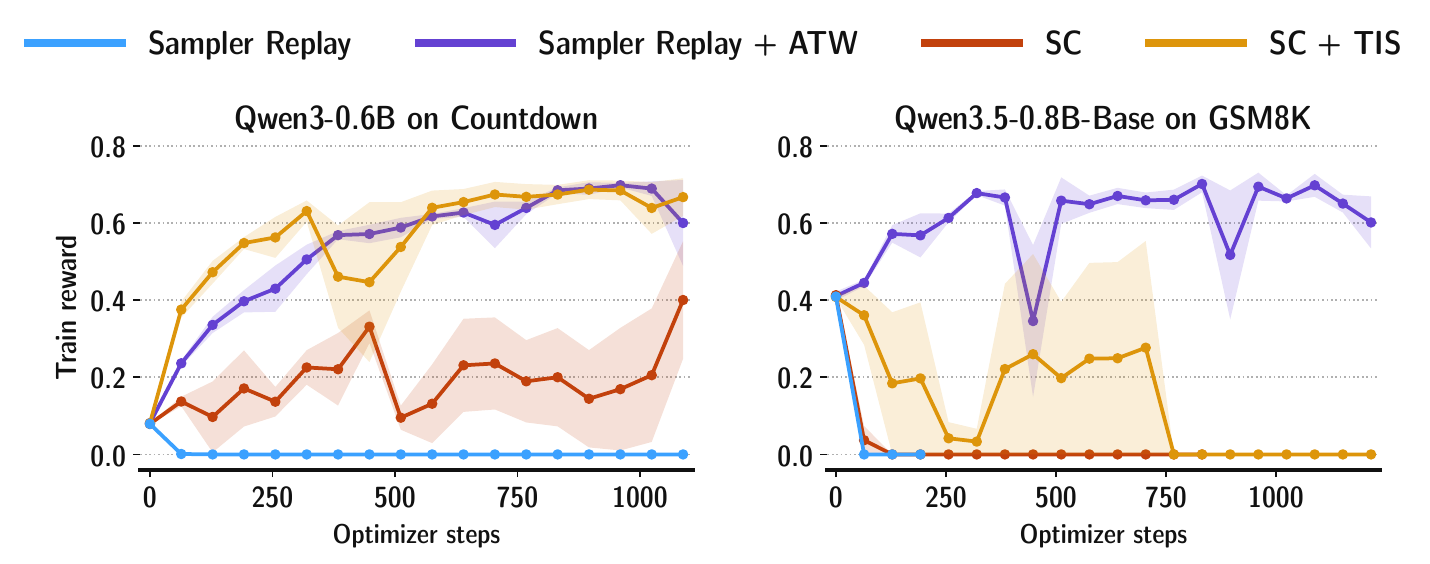}
\caption{Qwen3-0.6B on Countdown and Qwen3.5-0.8B-Base on GSM8K, sampler syncs weights every 64 step. Sampler Replay and SC fail. Sampler Replay + ATW($\tau=1e-4$) and SC+TIS train on Qwen3-0.6B. Only Sampler Replay+ATW($\tau=1e-3$) trains GSM8K. }
\label{fig:staleness}
\end{figure}

\subsection{Ablating Replays}
We demonstrate the contribution of replaying different parts of the model by ablating Sampler Replay. We reproduce our Countdown INT8 and noise setup from \Cref{sec:small_scale} and compare replaying logits, hidden states, or both. As shown in \Cref{fig:ablation} both Logit Replay and Hidden State Replay help, but combining them is more effective. This implies that more replay is generally better. In \Cref{sec:score_centering_equals_logits_replay}, we show that, mathematically, Logits Replay equals Score Centering. In our experiments here, we note that Score Centering and Logits Replay have empirically similar training curves, further strengthening this point. 

\begin{figure}[ht!]
\centering
\includegraphics[scale=0.57]{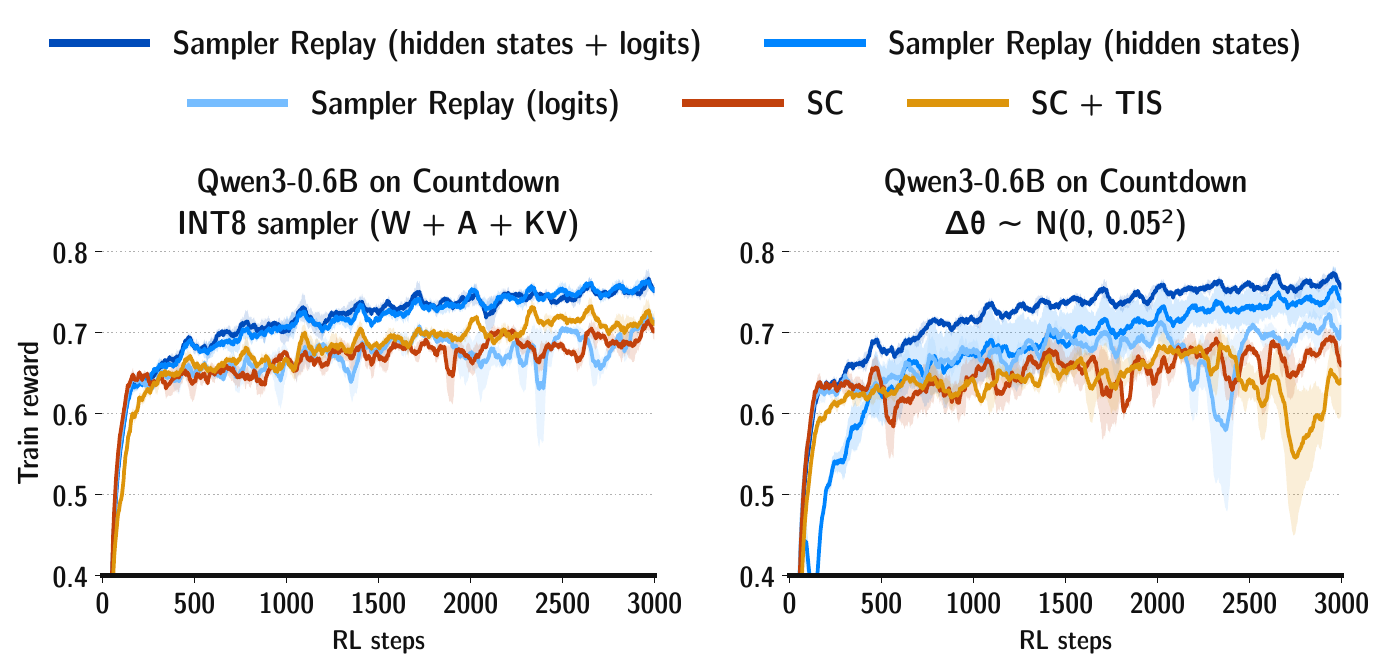}
\caption{Ablating the replays: Qwen3-0.6B on Countdown with an INT8 sampler and with sampler weight noise, replaying the sampler's hidden states and logits, its hidden states only or its logits only, against SC and SC + TIS. As proved, replaying logits shows similar training as SC. The benefits of replaying logits and hidden states compound.}
\label{fig:ablation}
\end{figure}

\subsection{KL Regularization Toward the Sampler}
\label{sec:sampler_kl}

\begin{figure}[h]
\centering
\includegraphics[scale=0.47]{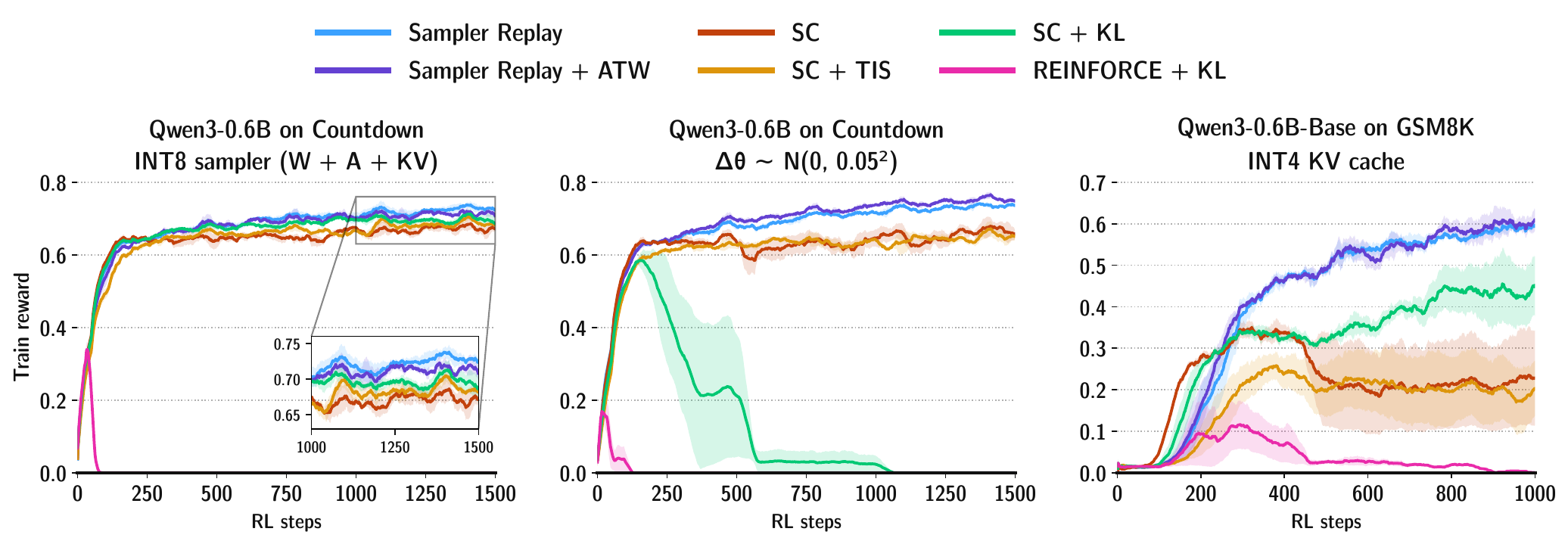}
\caption{Sampler Replay outperforms KL regularization toward the sampler. Experiments in
\Cref{fig:small_scale} setting, with $\beta=0.01$ on Countdown and $\beta=0.0001$
on GSM8K. On Countdown, SC + KL slightly improves on SC with the INT8
sampler and collapses under weight perturbation. REINFORCE + KL collapses
in both Countdown settings.}
\label{fig:sampler_kl}
\end{figure}

We now compare Sampler Replay to a recent method for matching the trainer and sampler: KL regularization. Following \citet{tml2026stablereinforce}, we add an auxiliary KL penalty loss toward the sampler on top of the trainer's policy-gradient loss. For each response token $y_t$ drawn from the sampler $\mu$, let $r_t=\pi_\theta(y_t\mid x,y_{<t})/\mu(y_t\mid x,y_{<t})$. We estimate $\mathrm{KL}(\mu\|\pi_\theta)$ using $k_3(r_t)=r_t-1-\log r_t$ and add its mean over response tokens to the loss with coefficient $\beta$. We clip $\log r_t$ to $[-20,20]$ and cap the per-token penalty at $10$.

As shown in \Cref{fig:sampler_kl}, adding the KL penalty to Score Centering \citep{tml2026stablereinforce} modestly improves SC on Countdown with
the INT8 sampler, but collapses under weight
perturbation. Vanilla REINFORCE + KL collapses in all settings. On GSM8K, using the standard $\beta=0.01$ for SC + KL causes training to stay near zero reward across all three seeds. We therefore search across $\beta$ and find the optimal
$\beta = 10^{-4}$ improves final training reward over SC, though it
remains below Sampler Replay.

\section{Estimating the Sampler's Gradient with Sampler Replay}
\label{sec:how_sampler_estimtates_sampler_gradient}
\input{math_section}

\subsection{Illustrating Trainer-Sampler Gradient Alignment}
\label{sec:weight_perturbation_and_sampler_replay}

Here, we use the weight perturbation setting of \Cref{sec:small_scale} to create a controlled setup that enables measuring exactly how closely the trainer's gradient approximates the sampler's. The sampler uses weights $\phi=\theta+0.05\,\epsilon\odot\theta_0$, where $\theta$ and $\theta_0$ denote the current and initial trainer weights, respectively, and $\epsilon\sim\mathcal{N}(0,I)$ is drawn once per run. At each step, we compute a reference gradient by differentiating the REINFORCE loss at $\phi$ on the same trajectories used for the method's update. We compare the direction and magnitude of the trainer's gradient with this reference in each decoder layer.

\begin{figure}[t]
\centering
\includegraphics[scale=0.47]{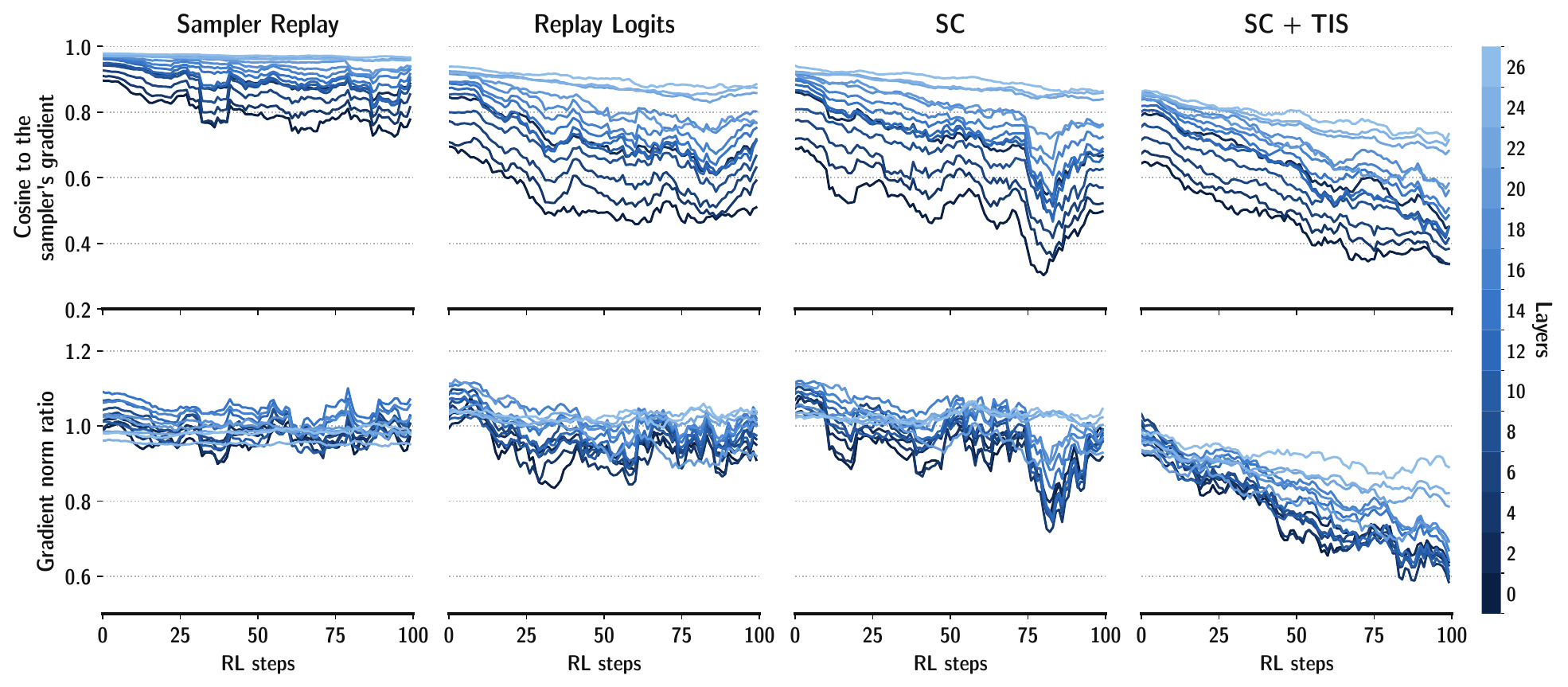}
\caption{Layerwise gradient alignment for Qwen3-0.6B on Countdown with
perturbed sampler weights (computed on 2{,}048 trajectories
each step). The top row shows cosine similarity between the trainer's
gradient and the sampler's. The bottom row shows the trainer's gradient
norm divided by the sampler's. Sampler Replay achieves high cosine similarity and the same norm as the sampler's true gradient using the trainer despite the weight offset. }
\label{fig:grad_check_layers}
\end{figure}

In \Cref{fig:grad_check_layers}, we find gradient alignment is weakest in the early layers and degrades over training, despite fixed weight perturbation. Replaying logits yields similar alignment to top-$k$ Score Centering but replaying hidden states improves alignment substantially. Over the final 20 steps, the mean cosine similarity in layer~0 is $0.76$ with Sampler Replay, compared with $0.49$ for logit replay alone, $0.43$ for SC, and $0.37$ for SC + TIS.

Sampler Replay also keeps the gradient norm ratio close to one across layers. With SC + TIS, the trainer's gradients become progressively smaller relative to the sampler's, particularly in the early layers. Over the same final 20 steps, the mean norm ratio in layer~0 is $0.65$ with SC + TIS and $0.97$ with Sampler Replay.

\subsection{Layerwise Train-Inference Mismatch}
\label{sec:per_layer_replay_effect}
We visualize the relative severity of train-inference mismatch across layers with different sampler quantizations with or without Sampler Replay in \Cref{fig:layer_gap}. Without Sampler Replay, as expected, the mismatch accumulates as the computations move through layers, but up to a point. Surprisingly, it reduces at final layers. It indicates close next-token distributions do not guarantee close forward passes. With Sampler Replay, the mismatch stops accumulating and drops to much lower values. This supports our hypothesis that Sampler Replay stops accumulation of mismatch across layers, keeping the forward passes close.

\begin{figure}[h!]
\centering
\includegraphics[scale=0.47]{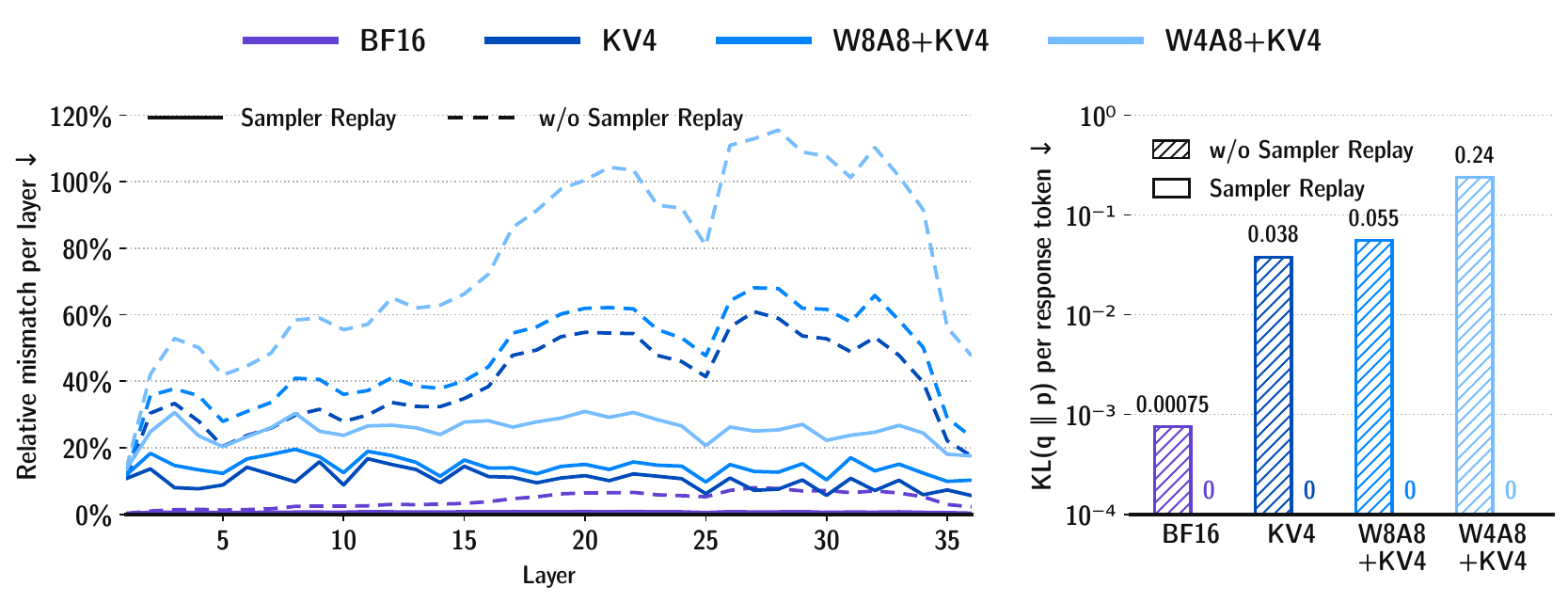}
\caption{Mismatch between the trainer's and the sampler's forward passes on Sokoban rollouts  on Qwen3-4B-Instruct-2507, with BF16 trainer, computed on 128 rollouts per sampler. Left: per layer, the hidden-state mismatch relative to the size of the sampler's residual update. with Sampler Replay (solid) and without Sampler Replay (dashed). With more severe quantization, the mismatch increases. Sampler Replay avoids accumulation of mismatches across layers. Right:  per token KL divergence between the sampler's ($q$) and the trainer's ($p$) next-token distributions without Sampler Replay (hatched); with Sampler Replay the trainer's logits equal the sampler's, so it is 0.}
\label{fig:layer_gap}
\end{figure}

\section{Sharp Bits of Sampler Replay Infrastructure}
\label{sec:sharp_bits_of_implementation}

We implement Sampler Replay in verl~\citep{sheng_hybridflow_2025}, leveraging vLLM~\citep{kwon2023vllm} for sampling and FSDP \citep{zhao_pytorch_2023} as the training engine. In this section, we first describe how storing sampler's hidden states in node-local storage enables overlapping transfer with computation to keep low end-to-end overhead. Next, we detail the quantization configurations used in rollout engines for different setups.

\subsection{Streaming Hidden-State Replay}
\label{sec:streaming_hidden_state_replay}
We record the BF16 residual stream entering each decoder layer and the
final normalization. A single rollout batch can produce hundreds of GiB
of hidden states, so we store them in node-local RAM or on local disk.
Inspired by how vLLM manages per-token state for PagedAttention, we use
two preallocated GPU buffers to stage the writes. While the sampler fills
one, the other drains asynchronously to local storage. This keeps GPU
staging memory fixed as rollouts grow. During training, we load the states
for microbatch $i+1$ while microbatch $i$ runs its forward and backward
passes. Appendix~\ref{app:sec:experiment_details} describes in more detail how the sampler records and stores the hidden states and how the trainer replays them.

When the sampler and trainer share an unquantized LM head, we avoid transferring full-vocabulary logits. We replay the head's input, the sampler's final hidden state \emph{after} normalization, instead. We show that it is the same as replaying logits. In synchronous Sokoban, the sampled-token KL estimate has the same numerical floor as full logit replay; smaller than $5\times10^{-8}$. On Qwen3-4B, this takes 5~KiB per token, compared with 297~KiB logits. We use this in our 4B DAPO and Sokoban experiments. In the other experiments we replay the logits themselves; in the INT8 Countdown and 30B experiments the sampler's LM head is quantized.

Overlapping transfers with computation keeps the wall-clock cost of replay small. In our Sokoban experiment(~\Cref{sec:sokoban}), a batch of 256 responses at the 8192-token limit can produce 370~GiB of hidden states. Even at this scale, Sampler Replay kept pace with Score Centering in end-to-end training (\Cref{fig:sokoban_walltime}).

\subsection{Quantized Rollout Engines}

The sampler's quantized weights track the trainer throughout RL. At each synchronization, we quantize the trainer's weights with symmetric round-to-nearest, using scales derived from the weights themselves. W8A8 uses INT8 weights with one scale per output channel. W4A8 uses INT4 weights with group-wise scales: each output channel has one scale per group of 128 input channels. Both samplers use INT8 activations with scales computed per token during rollout, so refreshing the sampler requires no calibration pass.

For the GSM8K INT4 KV runs and our 4B DAPO and Sokoban experiments, we use a packed INT4 cache with QuaRot's post-RoPE Hadamard rotation on queries and keys~\citep{ashkboos2024quarot}. As in the KV4 baseline of \citet{park2026kv4rl}, we encode rotated keys and unrotated values separately per token and KV head, with an FP16 scale and offset per vector. Because vLLM's \texttt{int4\_per\_token\_head} backend was slow, we used Claude to develop CUDA kernels for the packed INT4 cache that achieved roughly twice the output throughput of BF16 attention on H100 and A100. For 30B INTELLECT-2, we follow \citet{marek2026score} and store unrotated keys and values in BF16 after rounding them to a symmetric INT4 grid.

\section{Related Works}
Most approaches to TIM treat the sampler's rollouts as off-policy data for the trainer \citep{sutton_reinforcement_2018,espeholt_impala_2018}. 
Truncated Importance Sampling (TIS) \citep{yao2025offpolicy}, 
Masked Importance Sampling (MIS) \citep{liu-li-2025-rl-collapse},
and IcePop \citep{lingteam2025ring1t_icepop}
trade bias for variance reduction by truncating or masking importance ratios. 
GRPO \citep{shao2024deepseekmath} clips token-level ratios, 
GSPO \citep{zheng2025gspo} clips sequence-level ratios,
and DPPO \citep{qi2026stablerl} uses policy divergence to determine which tokens to mask.
These methods weight or mask gradient contributions to improve the trainer's policy. We instead seek to better estimate the sampler's gradient.

A complementary line of research aims to make the trainer and sampler more similar to each other. 
Learning approaches regularize the trainer towards the sampler with KL divergence \citep{tml2026stablereinforce,kimiteam2025kimik2}.
Infrastructure approaches try to make training and inference arithmetic bit-wise identical \citep{qi2025defeating,he2025nondeterminism,bai2026bitwise} and quantize the trainer the same way as the sampler \citep{gu2026qarl}.
But it is difficult to achieve and maintain complete bit-wise parity between trainer and sampler while also optimizing asynchronous RL throughput \citep{noukhovitch_asynchronous_2025}. There will likely remain some TIM, making these methods complementary to our approach. 

Sampler replay and reproducing the sampler's forward pass in the trainer are an effective paradigm encompassing multiple previous works on stabilizing RLVR. 
Keep Sampling Mask can be seen as replaying the top-p mask of the sampler~\citep{deepseek2025v32,vllm2026samplingmask}.
Rollout Routing Replay (R3) replays the sampler's expert routing decisions \citep{ma2025r3}. 
QUADS reproduces the sampler's weight quantization in the trainer \citep{zhuge2026quads}, while TRACE replays the sampler's quantization scales and rounding information \citep{wang2026trace}.
Furthermore, we show that full-vocabulary Score Centering \citep{marek2026score} is mathematically equivalent to replaying the sampler's logits.
Our framing offers a common explanation for the effectiveness of these methods: by intervening in the trainer's forward pass to follow sampler's, they can better approximate the sampler's gradient.

\section{Limitations}
\label{sec:limitations}
Activation Trust Weighting, though empirically quite effective, does not have strong theoretical grounding. Optimization theory may offer more principled ways to estimate the sampler's gradient.
Also, though we have tested Sampler Replay in policy-lag and data staleness setups, we have not explicitly implemented it in asynchronous RLVR \citep{noukhovitch_asynchronous_2025}. Various async RLVR adoptions of Score Centering imply that Sampler Replay is likely to be effective there as well, though it may require further investigation of design choices. For example, in async RLVR, a single rollout may be generated over multiple updates of the sampler \citep{piche_pipelinerl_2025}. Therefore, we can either replay the hidden states recorded during rollout or recompute them with the current weights. The former is easier to implement, but it remains unclear which approach achieves better performance.

\section{Conclusion}
Current approaches view TIM as a distribution mismatch in which the sampler's rollouts used to estimate the trainer's gradient are off-policy. In short, they train the trainer using an imperfect sampler. We propose a novel framing that the sampler is the final product and should be the target of our gradient updates. We view TIM as an optimization problem in which the trainer inaccurately estimates the sampler's gradient. We propose Sampler Replay, an algorithm that replays the sampler's hidden states and logits in the trainer's forward pass to better approximate the sampler's backward pass. In short, we train the sampler using an imperfect trainer. We empirically show that Sampler Replay stabilizes RLVR across diverse settings with severe TIM, often outperforming Score Centering. Our framing also explains how several existing methods are effective at improving RLVR. In particular, we establish a mathematical equivalence and also empirical similarity between Score Centering and replaying the sampler's logits in the trainer. We hope our work inspires better algorithms for training the sampler, not the trainer.
\bibliographystyle{plainnat}
\bibliography{references}

\clearpage
\beginappendix

\section{Extra Experimental Details}
\label{app:sec:experiment_details}

\subsection{Data Flow of Streaming Hidden-State Replay}
\label{app:sec:streaming_replay}
\begin{figure}[h!]
\centering
\includegraphics[scale=0.47]{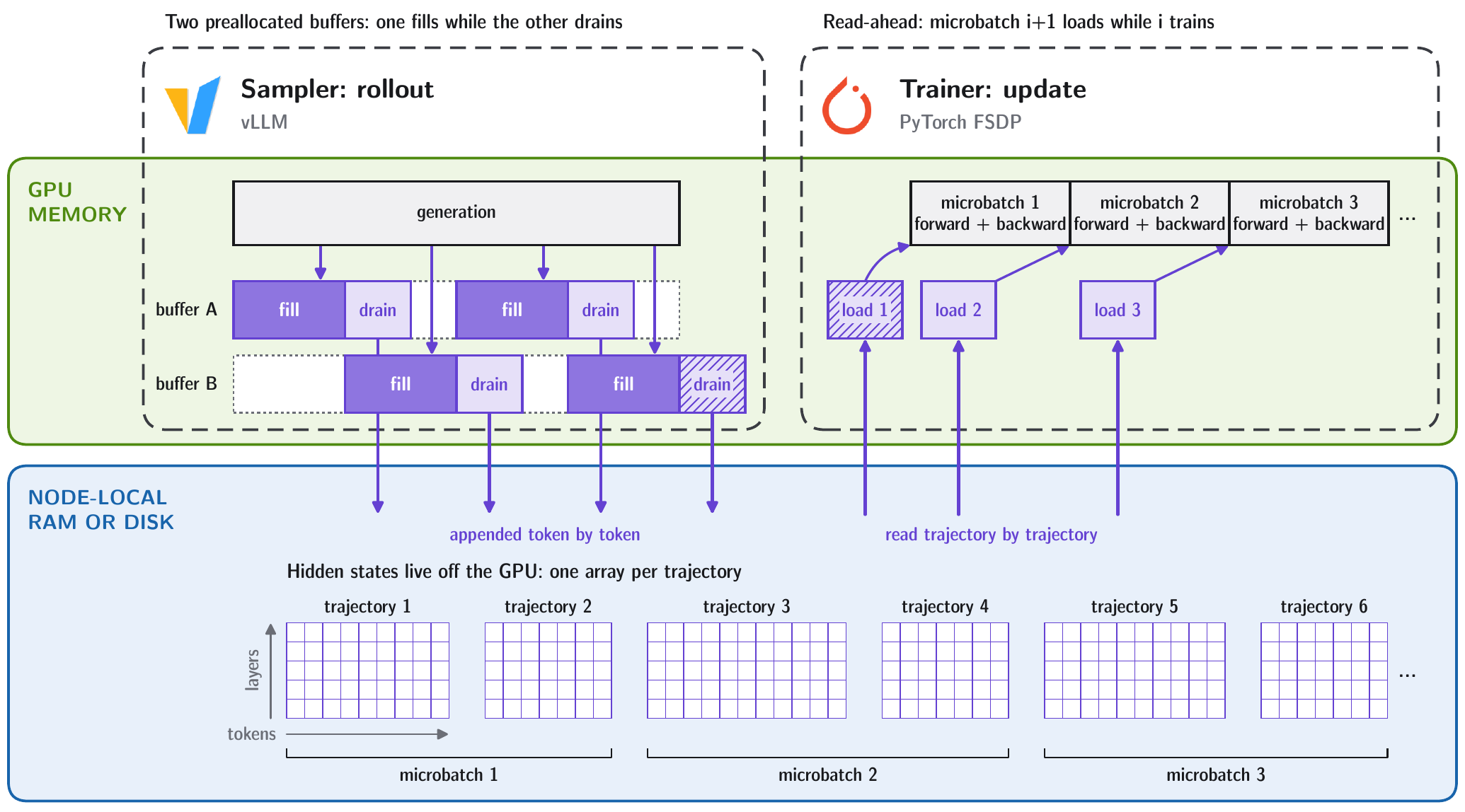}
\caption{Streaming hidden-state replay in one RL step (\Cref{sec:streaming_hidden_state_replay}). vLLM fills one GPU buffer while the other drains to node-local files grouped by trajectory. FSDP loads the next microbatch's states while training on the current one. Hatching marks the final drain and the first load.}
\label{fig:streaming_replay}
\end{figure}

This section expands on \Cref{sec:streaming_hidden_state_replay} with
implementation details of streaming hidden-state replay in verl.
\Cref{fig:streaming_replay} shows how the sampler's hidden states move
between GPU memory and node-local storage during one RL step.

We reuse the auxiliary hidden-state outputs that vLLM provides for EAGLE-3
speculative decoding. We enable them before compilation and CUDA graph
capture, so we obtain the states during the sampler's ordinary forward
pass. We record the embedding output and, after each decoder layer, the
BF16 sum of the residual and the layer update. These are the inputs to
every decoder layer and to the final normalization.  
We record these layer inputs while vLLM processes the prompt and generates
the response. If vLLM preempts a request and recomputes its prefix, we keep the states
from the original forward pass. Double buffering lets vLLM continue generating while background threads write the states in bulk, using one file per recorded quantity for each trajectory, with rows indexed by token position.

The FSDP trainer packs the trajectories in each microbatch into a single
sequence and reads their saved states in the same token order. While one
microbatch trains, it reads the next microbatch's states into pinned host
memory and copies them to the GPU on a separate CUDA stream. Forward hooks
replace the trainer's values with the sampler's recorded values at each
decoder layer, at the final normalization, and, for logit replay, at the
LM head. Gradients still flow through the trainer's own computation.
When gradient checkpointing recomputes a layer during backward, the hook
restores the same recorded input.

We use \texttt{/dev/shm} by default and node-local disk for long rollouts.
The sampler finishes its pending writes before the update. After the
update, we remove the trajectory files in the background and reuse the
disk blocks across steps. In a four-H100 run with Qwen3-4B-Base, 256
responses of 4{,}096 tokens produced 197~GiB of states per step.
Generation and update times on disk were each within one second of their
times in RAM across all three measured steps. At 8{,}192 tokens, the disk
held 387~GiB per step. In our Qwen3-4B Countdown profile on one H100
(\Cref{sec:streaming_hidden_state_replay}), recording and replaying
12.1~GB of hidden states added 0.76~s in total across generation and
the trainer's forward and backward passes. Our implementation uses one GPU per sampler replica on a single node.
We leave support for tensor-parallel samplers and multi-node replay to
future work.

\subsection{Wall-Clock Time on Sokoban}
\begin{figure}[h!]
\centering
\includegraphics[scale=0.57]{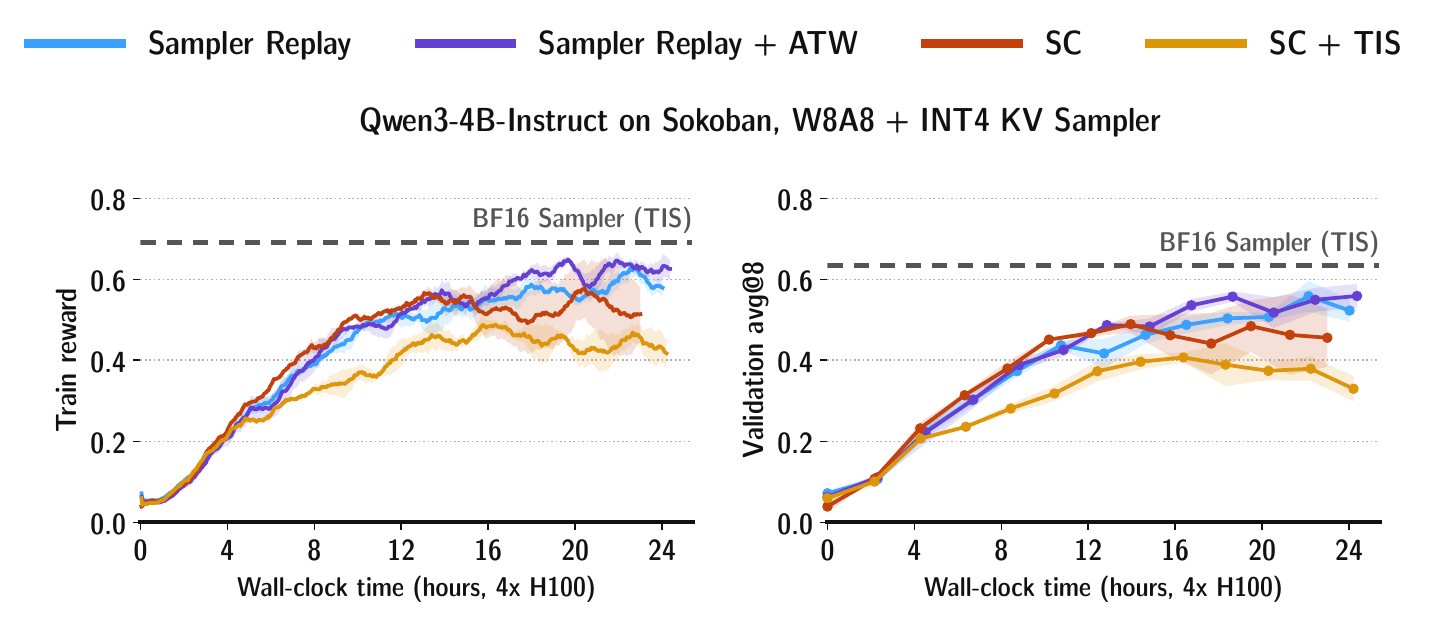}
\caption{Sampler Replay trains with negligible overhead. Qwen3-4B-Instruct on Sokoban (\Cref{fig:sokoban}) against wall-clock hours on 4 H100s, up to step 600. Wall-clock time sums the training steps and the validation passes; Dashed grey: TIS on an unquantized BF16 sampler at step 600.}
\label{fig:sokoban_walltime}
\end{figure}

\subsection{Activation Trust Weighting on Score Centering}
\label{sec:sc_match}
\begin{figure}[h!]
\centering
\includegraphics[scale=0.57]{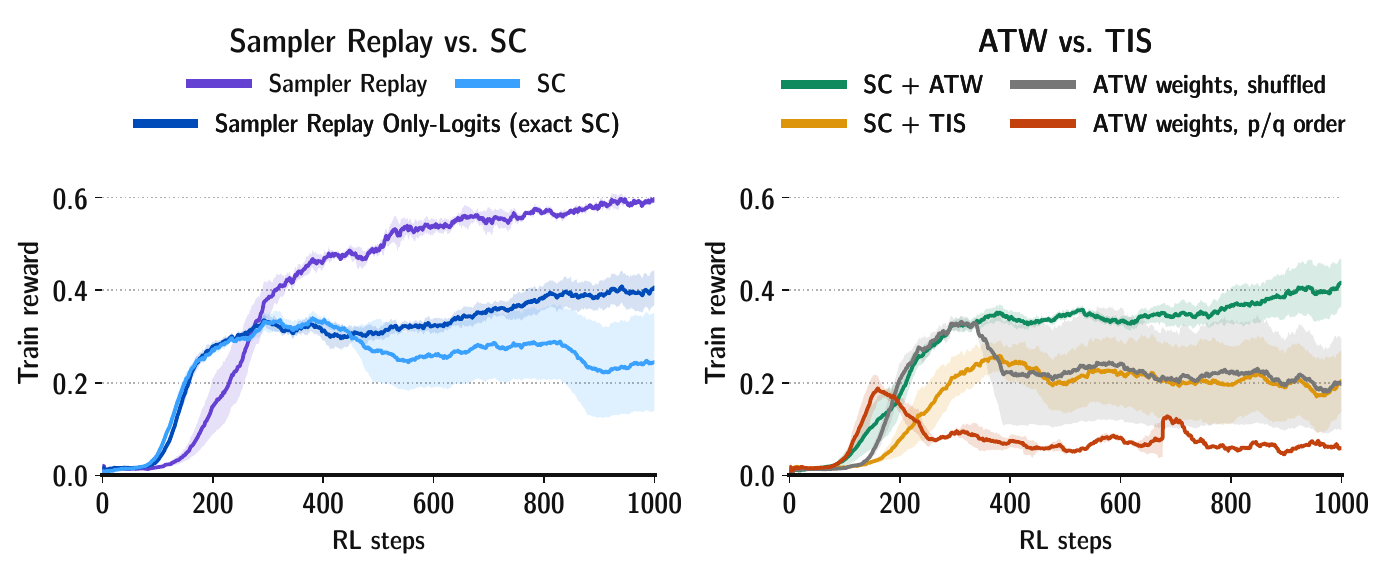}
\caption{Qwen3-0.6B-Base on GSM8K with an INT4 KV cache. Left, replaying
hidden states and logits outperforms logit replay alone (exact SC). This shows replaying hidden states on top of logits matters.
Right, ATW improves even SC. Reassigning the same weights at random or by
importance ratio lowers reward. It shows the ATW's specific weight assignment to tokens matters as reshuffling the same weights harms training.}
\label{fig:sc_match}
\end{figure}

We compare Score Centering and Sampler Replay on GSM8K with Qwen3-0.6B-Base
and an INT4 KV cache. Replaying only the sampler's logits gives the same
gradient as SC over the full vocabulary
(\Cref{sec:score_centering_equals_logits_replay}). At step 1{,}000, this
exact form of SC reaches a training reward of about $0.40$, compared with
$0.24$ for top-128 SC. Replaying the sampler's hidden states as well raises
reward to $0.60$ (\Cref{fig:sc_match}, left).

ATW also improves top-128 SC. We run the trainer on its own hidden states
and compare each layer's output with the sampler's recorded output. Tokens
with a larger mismatch at any layer receive a smaller weight
(\Cref{sec:atw}, $\tau=0.5$). Adding ATW raises the final training reward
of SC from about $0.24$ to $0.41$.

The benefit of ATW depends on which tokens it downweights. We keep the
same weights and change only which tokens receive them. In \emph{shuffled},
we assign the weights at random. In \emph{$p/q$ order}, we give the smallest
weights to the tokens with the smallest importance ratios. Both assignments
yield lower reward than ATW, which gives the smallest weights to tokens
with the largest hidden-state mismatch (\Cref{fig:sc_match}, right).

\subsection{Accounting for mismatch through ATW vs TIS}
\label{app:atw_vs_tis}

\paragraph{Why is importance sampling not effective in Sampler Replay?}
Sampler Replay tries to reproduce the sampler's forward pass in the trainer, but they can differ significantly for some tokens. 
This mismatch is due to a variety of factors including differing kernels~\citep{he2025nondeterminism} and data ~\citep{espeholt_impala_2018}. Empirically, we find that many tokens have high mismatch when the trainer is many updates ahead of the sampler that generated the data i.e. staleness. Previous approaches have handled this mismatch with Truncated Importance Sampling \citep[TIS;][]{yao2025offpolicy} which reweighs the gradient of each token by the ratio of probabilities between trainer and sampler.
\begin{equation}
g_\pi = \mathbb{E}_{x \sim D, y \sim \mu} 
\sum_{t=1}^{|y|} A_t
\underbrace{
    \max(\frac{\pi(y_t)}{\mu(y_t)}, C)
}_{\text{Truncated Importance Sampling}} 
\gradtheta \pi_\theta(y_t\mid x,y_{<t})
\end{equation}
where $C$ is a truncation hyperparameter.
But replaying the logits, as done in Score Centering and Sampler Replay, interferes with this method. If we use the sampler's logits in calculating the trainer's probability $\pi(y_t) = \sigma( W_{\text{head}} h_{\text{logit}}^\mu(y_t))$, the two methods can only differ in their LM head calculation. Given recent choices of FP32 LM head \citep{chen2025minimax}, the importance sampling ratio becomes nearly always 1, and does not benefit training. If we instead treat the trainer's forward pass as being unrelated to the sampler $\pi(y_t) = \sigma( W_{\text{head}} h_{\text{logit}}^\sigma(y_t))$, following the implementation of Score Centering \citep{marek2026score}, then we do not accurately estimate the trainer forward pass' true probability over $p(y_t)$. Empirically, we find this can negatively impact performance, see \Cref{sec:small_scale}. This motivates ATW as an alternative approach to down-weighing token-level gradients without importance sampling.

\begin{figure}[h!]
\centering
\includegraphics[scale=0.47]{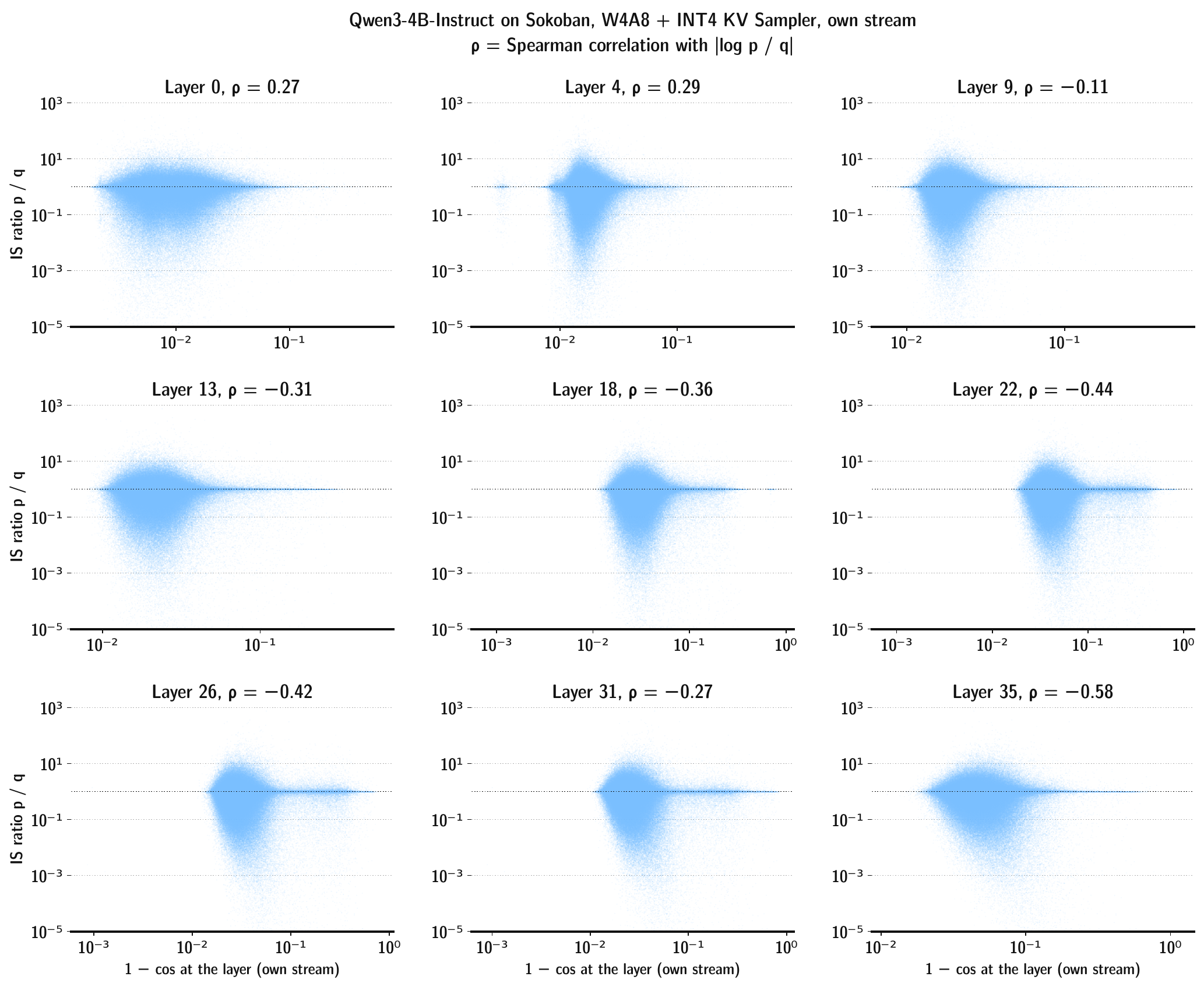}
\caption{Hidden-state mismatch and importance ratios do not correlate strongly. Measured on 128 Sokoban rollouts
from Qwen3-4B-Instruct with a W4A8 + INT4 KV sampler. Each panel compares
the sampled token's importance ratio $p/q$ with the cosine distance between
sampler and trainer outputs at one layer, measured without replay. Even
tokens with $p/q$ near one span a wide range of hidden-state mismatch.
$\rho$ is the Spearman correlation between cosine distance and $|\log(p/q)|$.}
\label{fig:is_vs_cos}
\end{figure}

\paragraph{How strongly do ATW weights correlate with importance sampling weights on the same response tokens?} We use the
Sokoban rollouts from \Cref{fig:layer_gap}, generated by Qwen3-4B-Instruct
with W4A8 weights and activations and an INT4 KV cache. We run the BF16
trainer on the same token sequences without replay. For each sampled
response token, we compute the importance ratio $p/q$, where $p$ and $q$
are the probabilities assigned to that token by the trainer and sampler.
At the position that predicts the token, we also measure the cosine
distance between the trainer's and sampler's outputs at each decoder layer.

We then apply both weighting rules to the same tokens. ATW uses the
largest cosine distance across all 36 layers with $\tau=0.1$, and
truncated IS uses $\min(p/q,2)$. The two weights have a Spearman
correlation of $-0.05$. Of the tokens that ATW gives zero weight,
$80.5\%$ receive an importance weight of at least one. On these rollouts,
the two methods downweight different tokens.

\input{hparams}

\end{document}

%% file: math_commands.tex
\usepackage{amsmath,amsfonts,bm}

\def\eqref#1{equation~\ref{#1}}

\def\1{\bm{1}}

\def\rb{{\textnormal{b}}}

\DeclareMathAlphabet{\mathsfit}{\encodingdefault}{\sfdefault}{m}{sl}
\SetMathAlphabet{\mathsfit}{bold}{\encodingdefault}{\sfdefault}{bx}{n}

%% file: assets/figstyles.tex
\usepackage[most]{tcolorbox}
\usepackage{tikz}
\usetikzlibrary{arrows.meta,positioning}

\definecolor{bfPrompt}{HTML}{6F42C1}
\definecolor{bfGood}{HTML}{1E9E6A}
\definecolor{bfForce}{HTML}{D9534F}

\newtcbox{\cue}{on line,
  colframe=bfForce, colback=bfForce!8,
  arc=1mm, boxrule=0.6pt,
  left=1pt, right=1pt, top=0pt, bottom=0pt}

\tcbset{
  bf-common/.style={enhanced, breakable, arc=2mm, boxrule=0.9pt,
    left=3mm, right=3mm, top=2mm, bottom=2mm, %
  },
  bf-prompt/.style={bf-common, colframe=bfPrompt, colback=bfPrompt!6, title=Prompt},
  bf-ok/.style={bf-common, colframe=bfGood, colback=bfGood!6, title=Response},
  bf-force/.style={bf-common, colframe=bfForce, colback=bfForce!5},
}

%% file: sampler_replay.tex
\providecommand{\srreplayy}{1cm}
\begin{tikzpicture}[x=1cm,y=\srreplayy]

  \definecolor{replay}{HTML}{6441D2}
  \definecolor{ink}{HTML}{18191C}

  \tikzset{
    module/.style={
      draw=ink,
      line width=0.85pt,
      rounded corners=3pt,
      minimum width=1.50cm,
      minimum height=0.72cm,
      fill=white,
      font=\sffamily\small,
      align=center
    },
    flow/.style={
      draw=ink,
      -{Latex[length=2.5mm,width=1.7mm]},
      line width=0.8pt
    },
    replayline/.style={
      draw=replay,
      dash pattern=on 4pt off 3pt,
      line width=1.15pt
    },
    replaydot/.style={
      circle,
      fill=replay,
      inner sep=0pt,
      minimum size=5.5pt
    },
    replaybox/.style={
      rectangle,
      rounded corners=5pt,
      draw=replay,
      line width=1.15pt,
      fill=white,
      inner xsep=8pt,
      inner ysep=8pt,
      font=\large,
      text=ink
    },
    replaytitle/.style={
      font=\sffamily\bfseries\small,
      text=ink
    },
    network/.style={
      draw=ink!45,
      fill=ink!4,
      line width=0.85pt,
      rounded corners=8pt
    },
    networklabel/.style={
      font=\bfseries\large
    },
    figtitle/.style={
      font=\bfseries\LARGE,
      text=ink
    }
  }


  \node[figtitle] at (4.65,7.65)
    {Sampler Replay};

  \begin{scope}[on background layer]
    \draw[network] (-1.65,-0.30) rectangle (1.65,7.59);
    \draw[network] (7.65,-0.30) rectangle (10.95,7.59);
  \end{scope}

  \node[networklabel, rotate=90, anchor=south]
    at (-1.67,3.645) {Sampler $\mu$};

  \node[networklabel, rotate=-90, anchor=south]
    at (10.97,3.645) {Trainer $\pi$};

  \node[module] (s1) at (0,0.36)
    {Layer 1};
  \node[module] (sL) at (0,3.17)
    {Layer $L$};
  \node[module] (sh) at (0,4.89)
    {LM head};
  \node[module] (ss) at (0,6.93)
    {softmax};

  \node[module] (t1) at (9.3,0.36)
    {Layer 1};
  \node[module] (tL) at (9.3,3.17)
    {Layer $L$};
  \node[module] (th) at (9.3,4.89)
    {LM head};
  \node[module] (ts) at (9.3,6.93)
    {softmax};

  \node (sx) at (0,-0.95) {$x$};
  \node (tx) at (9.3,-0.95) {$x$};

  \node (sp) at (0,8.13)
    {$\log p_\mu$};
  \node (tp) at (9.3,8.13)
    {$\log p_\pi$};

  \foreach \xx/\nn in {0/s,9.3/t}{
    \draw[flow] (\nn x.north) -- (\nn 1.south);
    \draw[flow] (\nn 1.north) -- (\xx,1.59);
    \node at (\xx,2.01) {$\vdots$};
    \draw[flow] (\xx,2.39) -- (\nn L.south);
    \draw[flow] (\nn L.north) -- (\nn h.south);
    \draw[flow] (\nn h.north) -- (\xx,5.70);
    \draw[flow] (\xx,6.12) -- (\nn s.south);
    \draw[flow] (\nn s.north) -- (\nn p.south);
  }

  \node[anchor=east] at (-0.12,1.17) {$h_\mu^1$};
  \node[anchor=east] at (-0.12,3.98) {$h_\mu^L$};
  \node[anchor=east] at (-0.12,5.87) {logits $\ell_\mu$};

  \node[anchor=west] at (9.42,1.17) {$h_\pi^1$};
  \node[anchor=west] at (9.42,3.98) {$h_\pi^L$};
  \node[anchor=west] at (9.42,5.87) {logits $\ell_\pi$};

  \foreach \yy in {1.17,3.98,5.87}{
    \draw[replayline] (0,\yy) -- (9.3,\yy);
    \node[replaydot] at (0,\yy) {};
    \node[replaydot] at (9.3,\yy) {};
  }

  \node[replaybox] (rb1) at (4.65,1.17)
    {$\displaystyle
      h_\pi^1 \leftarrow h_\pi^1
      +\operatorname{sg}(h_\mu^1-h_\pi^1)
    $};

  \node[replaybox] (rbL) at (4.65,3.98)
    {$\displaystyle
      h_\pi^L \leftarrow h_\pi^L
      +\operatorname{sg}(h_\mu^L-h_\pi^L)
    $};

  \node[replaybox] (rbl) at (4.65,5.87)
    {$\displaystyle
      \ell_\pi \leftarrow \ell_\pi
      +\operatorname{sg}(\ell_\mu-\ell_\pi)
    $};

  \node at (4.65,2.75) {$\vdots$};

  \foreach \rb/\yy in {rb1/1.17,rbL/3.98,rbl/5.87}{
    \fill[replay] ($(\rb.east)!0.5!(7.65,\yy)$) ++(0.12,0) -- ++(-0.24,0.10) -- ++(0,-0.20) -- cycle;
  }

  \node[replaytitle] at (4.65,1.92)
    {Hidden States Replay};

  \node[replaytitle] at (4.65,4.73)
    {Hidden States Replay};

  \node[replaytitle] at (4.65,6.62)
    {Logits Replay (Score Centering)};

\end{tikzpicture}%

%% file: math_section.tex
We examine how the trainer estimates the sampler's gradient with and without Sampler Replay, and why we hypothesize that Sampler Replay improves this estimate.

Both the sampler and the trainer have a sequence of $T$ layers. Let $\mu_i$ and $\pi_i$ denote the operations of the $i$-th layer of the sampler and the trainer, respectively, and $w_i^\mu$ and $w_i^\pi$ their corresponding weights. Without replay, the hidden states of the sampler and the trainer are computed as
\begin{equation}
\begin{aligned}
    h_i^\mu &= \mu_i(h_{i-1}^\mu), \\
    h_i^\pi &= \pi_i(h_{i-1}^\pi).
\end{aligned}
\end{equation}

With Sampler Replay, the trainer's hidden states are instead computed as
\begin{equation}
    \tilde{h}_i =
    \pi_i(\tilde{h}_{i-1})
    + \sg\left(h_i^\mu - \pi_i(\tilde{h}_{i-1})\right).
\end{equation}

We want to estimate the gradient of the loss $\mathcal{L}$ with respect to the sampler's weights, $\nabla_\phi \mathcal{L}$, where $\phi = \{w_i^\mu\}_{i=1}^T$. Since the inference engine does not support backpropagation, we use the trainer to estimate this gradient.

Let us focus on a single layer $i$. Write $J_i^\mu(h)$ and $J_i^\pi(h)$ for its weight Jacobians at input $h$, and let $g_i$ and $\delta_i$ denote its weight gradient and upstream gradient, respectively. The superscripts $\mu$, $\pi$, and $\mathrm{replay}$ on these gradients identify the backward pass. Both $g_i^\pi$ and $g_i^{\mathrm{replay}}$ are computed with respect to $w_i^\pi$. By the chain rule, the gradient we seek to estimate is
\begin{equation}
    \underbrace{g_i^\mu}_{\text{Target gradient}}
    =
    \underbrace{J_i^\mu(h_{i-1}^\mu)^\top}_{\text{Layer Jacobian}}
    \underbrace{\delta_i^\mu}_{\text{Upstream gradient}}.
\end{equation}

Without Sampler Replay, the trainer estimates this gradient as
\begin{equation}
    g_i^\pi = J_i^\pi(h_{i-1}^\pi)^\top \delta_i^\pi.
    \label{eq:grad_p}
\end{equation}

With Sampler Replay, we instead estimate this gradient as
\begin{equation}
    g_i^{\mathrm{replay}}
    = J_i^\pi(h_{i-1}^\mu)^\top \delta_i^{\mathrm{replay}}.
    \label{eq:grad_q}
\end{equation}

The linear output head gives a case in which Eq.~\ref{eq:grad_q} is exact. Treat the head as layer $T$ and sampler logits $z_{\mathrm{sampler}}=w_T^\mu h_{T-1}^\mu$. For a sampled token $y$ with advantage $A$, let $p$ and $q$ denote the trainer's and sampler's next-token distributions, respectively. The sampler's loss $\mathcal{L}=-A\log q_y$ gives the following gradient with respect to the sampler head matrix:
\begin{equation}
    g_T^\mu = -A(e_y-q)(h_{T-1}^\mu)^\top.
\end{equation}
Without replay, Eq.~\ref{eq:grad_p} uses $p$ in place of $q$ and $h_{T-1}^\pi$ in place of $h_{T-1}^\mu$. By Theorem 1, replaying the logits recovers the sampler's loss gradient at the logits, $\delta_T^{\mathrm{replay}}=\delta_T^\mu=-A(e_y-q)$. Replaying the input to the head recovers the other factor, $(h_{T-1}^\mu)^\top$. Together, the two replays therefore give $g_T^{\mathrm{replay}}=g_T^\mu$.

Note that for earlier layers the trainer's Jacobian can still differ from the sampler's at that input. However, Sampler Replay removes one source of discrepancy: the hidden states at which the Jacobians are evaluated. Sampler Replay syncs the trainer with the sampler at each layer preventing accumulation of mismatch across layers. We hypothesize that removing this source of discrepancy yields a better approximation of the sampler's gradient. We examine this hypothesis empirically in \Cref{sec:weight_perturbation_and_sampler_replay} by comparing the gradients estimated by the trainer with and without Sampler Replay against the sampler's gradient in the weight perturbation setting. We empirically show that Sampler Replay avoids accumulation of mismatch in \Cref{sec:per_layer_replay_effect}.

%% file: hparams.tex
\section{Hyperparameters}
\label{app:sec:hyperparameters}


\begin{table}[h!]
\centering
\small
\caption{Qwen3-0.6B on Countdown, INT8 sampler.}
\label{tab:hp_countdown_int8}
\begin{tabular*}{0.85\linewidth}{@{\extracolsep{\fill}}lr@{}}
\toprule
Hyperparameter & Value \\
\midrule
Model & Qwen3-0.6B \\
Dataset & Countdown (generated) \\
Reward & 0 / 1 \\
Chat template & Qwen3, thinking off \\
Max sequence length & 512 (prompt + response) \\
Prompts per step & 64 \\
Responses per prompt & 8 \\
Sampling & temperature 1.0, no top-p / top-k \\
Optimizer & SGD, lr 0.01, weight decay 0, no clipping \\
Policy loss & REINFORCE \\
Loss aggregation & token mean \\
Advantage & $r - \text{mean}_{\text{group}}(r)$ \\
Weights & INT8, per channel \\
Activations & INT8, per token \\
Quantized layers & all linears + embedding + LM head \\
KV cache & INT8, per token-head \\
Training steps & 3{,}000 \\
\midrule
Algorithm & Setting \\
\midrule
Sampler Replay & hidden states + logits \\
Sampler Replay + ATW & hidden states + logits, $\tau=0.1$ \\
SC & top-128 \\
SC + TIS & top-128, TIS cap 2 \\
TIS & TIS cap 2 \\
\bottomrule
\end{tabular*}
\end{table}

\begin{table}[h!]
\centering
\small
\caption{Qwen3-0.6B on Countdown, sampler weight perturbation.}
\label{tab:hp_countdown_noise}
\begin{tabular*}{0.85\linewidth}{@{\extracolsep{\fill}}lr@{}}
\toprule
Hyperparameter & Value \\
\midrule
Model & Qwen3-0.6B \\
Dataset & Countdown (generated) \\
Reward & 0 / 1 \\
Chat template & Qwen3, thinking off \\
Max sequence length & 512 (prompt + response) \\
Prompts per step & 64 \\
Responses per prompt & 8 \\
Sampling & temperature 1.0, no top-p / top-k \\
Optimizer & SGD, lr 0.01, weight decay 0, no clipping \\
Policy loss & REINFORCE \\
Loss aggregation & token mean \\
Advantage & $r - \text{mean}_{\text{group}}(r)$ \\
Weights & BF16, $\phi=\theta+0.05\,\epsilon\odot\theta_0$ \\
Activations & BF16 \\
Quantized layers & none \\
KV cache & BF16 \\
Training steps & 3{,}000 \\
\midrule
Algorithm & Setting \\
\midrule
Sampler Replay & hidden states + logits \\
Sampler Replay + ATW & hidden states + logits, $\tau=0.1$ \\
SC & top-128 \\
SC + TIS & top-128, TIS cap 2 \\
TIS & TIS cap 2 \\
\bottomrule
\end{tabular*}
\end{table}

\begin{table}[h!]
\centering
\small
\caption{Qwen3-0.6B-Base on GSM8K, INT4 KV sampler.}
\label{tab:hp_gsm8k_kv4}
\begin{tabular*}{0.85\linewidth}{@{\extracolsep{\fill}}lr@{}}
\toprule
Hyperparameter & Value \\
\midrule
Model & Qwen3-0.6B-Base \\
Dataset & GSM8K \\
Reward & 0 / 1 \\
Chat template & Qwen3 \\
Max prompt / response length & 1{,}024 / 4{,}096 \\
Prompts per step & 4 \\
Responses per prompt & 16 \\
Sampling & temperature 1.0, no top-p / top-k \\
Optimizer & AdamW ($\beta=(0.9, 0.999)$), lr 1e-6, weight decay 0, clip 1.0 \\
Policy loss & REINFORCE \\
Loss aggregation & token mean \\
Advantage & $r - \text{mean}_{\text{group}}(r)$ \\
Weights & BF16 \\
Activations & BF16 \\
Quantized layers & none \\
KV cache & INT4, affine, per token-head, Hadamard-rotated K \\
Training steps & 1{,}000 \\
Evaluation & GSM8K-Platinum hard 100, avg@32 \\
\midrule
Algorithm & Setting \\
\midrule
Sampler Replay & hidden states + logits \\
Sampler Replay + ATW & hidden states + logits, $\tau=1.0$ \\
SC & top-128 \\
SC + TIS & top-128, TIS cap 2 \\
TIS & TIS cap 2 \\
\bottomrule
\end{tabular*}
\end{table}


\begin{table}[h!]
\centering
\small
\caption{Qwen3-4B-Base on DAPO, W4A8 + INT4 KV sampler.}
\label{tab:hp_dapo_w4a8}
\begin{tabular*}{0.85\linewidth}{@{\extracolsep{\fill}}lr@{}}
\toprule
Hyperparameter & Value \\
\midrule
Model & Qwen3-4B-Base \\
Dataset & DAPO-Math-17k (English) \\
Reward & $\pm1$ \\
Chat template & Qwen3 \\
Max prompt / response length & 1{,}024 / 16{,}384 \\
Prompts per step & 16 \\
Responses per prompt & 16 \\
Sampling & temperature 1.0, no top-p / top-k \\
Optimizer & AdamW ($\beta=(0.9, 0.999)$), lr 1e-6, weight decay 0, clip 1.0 \\
Policy loss & REINFORCE \\
Loss aggregation & seq-mean-token-sum-norm ($\div$ 16{,}384) \\
Advantage & $r - \text{mean}_{\text{group}}(r)$ \\
Weights & INT4, group 128 \\
Activations & INT8, per token \\
Quantized layers & attention + MLP projections \\
KV cache & INT4, affine, per token-head, Hadamard-rotated K \\
Training steps & 500 \\
Evaluation & AIME 2024 / 2025, avg@32 \\
\midrule
Algorithm & Setting \\
\midrule
Sampler Replay & hidden states + head input (logits) \\
Sampler Replay + ATW & hidden states + head input (logits), $\tau=0.3$ \\
SC & top-128 \\
SC + TIS & top-128, TIS cap 2 \\
\bottomrule
\end{tabular*}
\end{table}

\begin{table}[h!]
\centering
\small
\caption{Qwen3-4B-Base on DAPO, W8A8 + INT4 KV sampler.}
\label{tab:hp_dapo_w8a8}
\begin{tabular*}{0.85\linewidth}{@{\extracolsep{\fill}}lr@{}}
\toprule
Hyperparameter & Value \\
\midrule
Model & Qwen3-4B-Base \\
Dataset & DAPO-Math-17k (English) \\
Reward & $\pm1$ \\
Chat template & Qwen3 \\
Max prompt / response length & 1{,}024 / 16{,}384 \\
Prompts per step & 16 \\
Responses per prompt & 16 \\
Sampling & temperature 1.0, no top-p / top-k \\
Optimizer & AdamW ($\beta=(0.9, 0.999)$), lr 1e-6, weight decay 0, clip 1.0 \\
Policy loss & REINFORCE \\
Loss aggregation & seq-mean-token-sum-norm ($\div$ 16{,}384) \\
Advantage & $r - \text{mean}_{\text{group}}(r)$ \\
Weights & INT8, per channel \\
Activations & INT8, per token \\
Quantized layers & attention + MLP projections \\
KV cache & INT4, affine, per token-head, Hadamard-rotated K \\
Training steps & 500 \\
Evaluation & AIME 2024 / 2025, avg@32 \\
\midrule
Algorithm & Setting \\
\midrule
Sampler Replay & hidden states + head input (logits) \\
Sampler Replay + ATW & hidden states + head input (logits), $\tau=0.3$ \\
SC & top-128 \\
SC + TIS & top-128, TIS cap 2 \\
\bottomrule
\end{tabular*}
\end{table}


\begin{table}[h!]
\centering
\small
\caption{Qwen3-4B-Instruct-2507 on Sokoban, W8A8 + INT4 KV sampler.}
\label{tab:hp_sokoban}
\begin{tabular*}{0.85\linewidth}{@{\extracolsep{\fill}}lr@{}}
\toprule
Hyperparameter & Value \\
\midrule
Model & Qwen3-4B-Instruct-2507 \\
Dataset & Reasoning Gym Sokoban (3 boxes) \\
Reward & 0 / 1 \\
Chat template & Qwen3-Instruct-2507 \\
Max prompt / response length & 512 / 8{,}192 \\
Prompts per step & 32 \\
Responses per prompt & 8 \\
Sampling & temperature 1.0, no top-p / top-k \\
Optimizer & AdamW ($\beta=(0.9, 0.999)$), lr 1e-6, weight decay 0, clip 1.0 \\
Policy loss & REINFORCE \\
Loss aggregation & seq-mean-token-sum-norm ($\div$ 8{,}192) \\
Advantage & $r - \text{mean}_{\text{group}}(r)$ \\
Weights & INT8, per channel \\
Activations & INT8, per token \\
Quantized layers & attention + MLP projections \\
KV cache & INT4, affine, per token-head, Hadamard-rotated K \\
Training steps & 600 \\
Evaluation & 128 held-out boards, avg@8 \\
\midrule
Algorithm & Setting \\
\midrule
Sampler Replay & hidden states + head input (logits) \\
Sampler Replay + ATW & hidden states + head input (logits), $\tau=0.1$ \\
SC & top-128 \\
SC + TIS & top-128, TIS cap 2 \\
\bottomrule
\end{tabular*}
\end{table}


\begin{table}[h!]
\centering
\small
\caption{Qwen3-30B-A3B-Base on INTELLECT-2 math, W8A8 + INT4 KV sampler.}
\label{tab:hp_moe_30b}
\begin{tabular*}{0.85\linewidth}{@{\extracolsep{\fill}}lr@{}}
\toprule
Hyperparameter & Value \\
\midrule
Model & Qwen3-30B-A3B-Base \\
Dataset & INTELLECT-2 math \\
Reward & -1 / 1 \\
Chat template & none \\
Max sequence length & 1{,}024 (prompt + response) \\
Prompts per step & 16 \\
Responses per prompt & 8 \\
Sampling & temperature 1.0, no top-p / top-k \\
Optimizer & SGD, lr 0.01, weight decay 0, no clipping \\
Policy loss & REINFORCE \\
Loss aggregation & token mean \\
Advantage & $r - \text{mean}_{\text{group}}(r)$ \\
Weights & INT8, per channel \\
Activations & INT8, per token \\
Quantized layers & \begin{tabular}[t]{@{}r@{}}all linears + experts + LM head; \\ embedding INT8 (weights only); \\ router BF16\end{tabular} \\
KV cache & INT4, symmetric, per token-head \\
Training steps & 800 \\
\midrule
Algorithm & Setting \\
\midrule
Sampler Replay & hidden states + logits \\
Sampler Replay + ATW & hidden states + logits + expert routing, $\tau=0.3$ \\
SC & top-128 \\
SC + TIS & top-128, TIS cap 2 \\
\bottomrule
\end{tabular*}
\end{table}


\begin{table}[h!]
\centering
\small
\caption{Staleness: Qwen3-0.6B on Countdown.}
\label{tab:hp_stale_countdown}
\begin{tabular*}{0.85\linewidth}{@{\extracolsep{\fill}}lr@{}}
\toprule
Hyperparameter & Value \\
\midrule
Model & Qwen3-0.6B \\
Dataset & Countdown (generated) \\
Reward & 0 / 1 \\
Chat template & Qwen3, thinking off \\
Max sequence length & 512 (prompt + response) \\
Prompts per step & 64 \\
Responses per prompt & 8 \\
Sampling & temperature 1.0, no top-p / top-k \\
Sampler weight sync & every 65 steps \\
Optimizer & SGD, lr 0.01, weight decay 0, no clipping \\
Policy loss & REINFORCE \\
Loss aggregation & token mean \\
Advantage & $r - \text{mean}_{\text{group}}(r)$ \\
Weights & BF16 \\
Activations & BF16 \\
Quantized layers & none \\
KV cache & BF16 \\
Training steps & 1{,}152 \\
\midrule
Algorithm & Setting \\
\midrule
Sampler Replay & hidden states \\
Sampler Replay + ATW & hidden states, $\tau=1e-4$ \\
SC & top-128 \\
SC + TIS & top-128, TIS cap 2 \\
\bottomrule
\end{tabular*}
\end{table}

\begin{table}[h!]
\centering
\small
\caption{Staleness: Qwen3.5-0.8B-Base on GSM8K.}
\label{tab:hp_stale_gsm8k}
\begin{tabular*}{0.85\linewidth}{@{\extracolsep{\fill}}lr@{}}
\toprule
Hyperparameter & Value \\
\midrule
Model & Qwen3.5-0.8B-Base \\
Dataset & GSM8K \\
Reward & $\pm1$ \\
Chat template & Qwen3.5, thinking off \\
Max prompt / response length & 2{,}048 / 4{,}096 \\
Prompts per step & 4 \\
Responses per prompt & 8 \\
Sampling & temperature 1.0, no top-p / top-k \\
Sampler weight sync & every 64 steps \\
Optimizer & SGD, lr 0.03, weight decay 0, no clipping \\
Policy loss & REINFORCE \\
Loss aggregation & seq-mean-token-sum-norm ($\div$ 4{,}096) \\
Advantage & $r - \text{mean}_{\text{group}}(r)$ \\
Weights & BF16 \\
Activations & BF16 \\
Quantized layers & none \\
KV cache & BF16 \\
Training steps & 1{,}280 \\
\midrule
Algorithm & Setting \\
\midrule
Sampler Replay & hidden states + head input \\
Sampler Replay + ATW & hidden states + head input, $\tau=1e-3$ \\
SC & top-32 \\
SC + TIS & top-32, TIS cap 2 \\
\bottomrule
\end{tabular*}
\end{table}